\documentclass[aps,pra,floatfix]{revtex4} 

\usepackage{graphicx}
\usepackage{dcolumn}
\usepackage{bm}

\usepackage{times}
\usepackage{graphicx}
\usepackage{subfigure}

\newcommand{\comment}[1]{}

\begin{document}

\title{Continuous Strategy  Replicator Dynamics for Multi--Agent Learning }

\author{Aram~Galstyan}
\affiliation{
Information Sciences Institute\\
University of Southern California\\
4676 Admiralty Way, Marina del Rey, CA 90292-6695\\
}

\date{\today}

\begin{abstract}
The problem of multi--agent learning and adaptation has attracted a great deal of attention in recent years. It has been suggested that the dynamics of multi agent learning can be studied using replicator equations from population biology. Most existing studies so far have been limited to discrete strategy spaces with a small number of available actions. In many cases, however, the choices available to agents are better characterized by continuous spectra. This paper suggests a generalization of the replicator framework that allows to study the adaptive dynamics of $Q$--learning agents with continuous strategy spaces.   Instead of probability vectors, agents' strategies are now characterized by probability measures over  continuous variables. As a result, the ordinary differential equations for the discrete case are replaced by a system of coupled integral--differential replicator equations that describe the mutual evolution of individual agent strategies. We derive a set of functional equations describing  the steady state  of the replicator dynamics, examine their solutions for several two--player games, and confirm our analytical results using simulations. 
\end{abstract}

\maketitle

\section{Introduction}
The notion of autonomous agents that learn by interacting with the environment, and possibly with other agents, is a central concept in modern distributed AI~\cite{SuttonBarto}. Of particular interests are systems where multiple agents learn concurrently and independently by interacting with each other.   This multi--agent learning problem has attracted a great deal of attention due to a number of important applications.   Among existing approaches,  multi--agent reinforcement learning (MARL) algorithms have become increasingly popular due to their generality~\cite{abdallah2008,Bowling02multiagentlearning,Claus1998,Hu98multiagentreinforcement,Peshkin00learningto,Hu2003,Stone2005}.  Although MARL does not hold the same convergence guarantees as in single--agent case,  it has been shown to work well in practice (for a  recent survey, see~\cite{Busoniu2008}).

From the analysis standpoint, MARL represents a complex  dynamical system, where the learning trajectories of individual agents are coupled with each other via  a collective reward  mechanism.  Thus, it is desirable to know what are the possible long--term behaviors of those trajectories. Specifically, one is usually interested whether, for a particular game structure, those trajectories converge to a    desirable steady state (called fixed points),    or oscillate indefinitely between many (possibly infinite) meta--stable states.  While answering this question has proven to be very difficult in the most general settings, there has been some limited progress  for specific scenarios. In particular, it has been established that for  simple, stateless Q--learning with a finite number of actions, the learning dynamics can be examined using the so called {\em replicator equations} from population biology~\cite{Hofbauer1988}. Namely, if one associates a particular biological trait with each pure strategy, then the adaptive learning of (possibly mixed) strategies in multi--agent settings is analogous to   competitive dynamics of mixed population, where the species evolve according to their relative fitness in the population.  This framework has been used successfully to study various interesting features of adaptive dynamics of learning agents~\cite{Sato2002,Sato2003,Tuyls2003,Tuyls2005,Panait2008,Vrancx2008,Hennes2009}.

With a few exceptions (e.g., see ~\cite{Ruijgrok2005,Tuyls2009}), most existing studies so far have focused on discrete action spaces, which has limited the full analysis of the learning dynamics to games with very few actions. On the other hand, in many practical scenarios, strategic interactions between agents are better characterized by continuous spectra of possible choices.  For instance, modeling an agent's bid in an auction with a continuous rather than discrete variable is more natural.  In such situations, agents' strategies are represented as  probability density functions defined over a continuous set of actions.  Of course, in reality all  the decisions are made over a discretized subset. However, the rationale for using the continuous approximation is that it makes the dynamics more amenable to mathematical analysis.

In this paper we consider simple $Q$--learning agents  that play repeated continuous--strategy games. The agents use Boltzmann action--selection mechanism that controls the exploration/exploitation tradeoff through a single temperature--like parameter. The reward functions for the agents are assumed to be functions of continuous variables instead of tensors, and the agent strategies are represented as probability distribution over those variables. In contrast to the finite strategy spaces where the learning dynamics is captured by a set of coupled ordinary differential equations, the replicator dynamics for  the continuous--strategy games are described by functional--differential equations for each agent, with coupling across different agents/equations. 

The long--term behavior of those equations define the steady state, or equilibrium, profiles of the agent strategies. It is shown that, in general, the steady state strategy profiles of the replicator dynamics  do not correspond to the Nash equilibria of the  game. This discrepancy can be attributed to the limited rationality of the agents due to exploration. In particular, for the Boltzmann action-selection mechanism studied here, exploration results in adding an entropic term to the agents' payoff function, with a coefficient  governed by the exploration rate (temperature). Thus, when one  decreases the exploration rate, the relative importance of this term diminishes and one is able to gradually recover the correspondence with the Nash equilibria. Furthermore, for games that allow {\em uniformly} mixed Nash equilibrium, the steady state solution of the replicator equation  is identical with this uniform Nash equilibrium for any  exploration rate. This is because the uniform distribution already maximizes the entropic  term. And example of such a game if provided in Section~\ref{sec:investment}. 

The rest of this paper is organized as follows: In the next section we provide a brief overview of relevant literature. In Section~\ref{sec:model} we introduce our model,  derive the replicator equations for the continuous strategy spaces, and a set of coupled non--linear functional equations that describe the steady state strategy profile. In Section~\ref{sec:example} we illustrate the framework on several examples of two--agent games, and  provide some detailed results for  general bi--linear and quadratic payoffs. Finally, we conclude the paper with a discussion of our results and possible future directions in Section~\ref{sec:discussion}.

\section{Background and Related Work}
\label{sec:back}

Reinforcement learning (RL)~\cite{SuttonBarto,Kaelbling} is a powerful
framework in which an agent  learns to behave optimally  through a trial and error exploration of the
environment.  At each step of interaction with the environment, the agent chooses an action based on the current state of the environment, and   receives a scalar reinforcement signal, or a reward, for that action. The agent's overall goal is to learn to act in a way that  will increase  the long--term cumulative reward.  Although RL was originally developed for single--agent learning in a stationary environment, it has been also generalized for multi--agent scenario. In the multi-agent setting,  the environment is highly dynamic because of the presence of other learning agents, and the
usual conditions for convergence to an optimal policy do not
necessarily hold~\cite{Claus1998,Kapetanakis2002,Shoham}.
Nevertheless,  various generalizations of single agent learning
algorithms  have been successfully applied to multi--agent
settings~\cite{abdallah2008,Bowling02multiagentlearning,Claus1998,Hu98multiagentreinforcement,Peshkin00learningto,Hu2003,Stone2005}.

Despite some empirical success, theoretical advances in multi--agent reinforcement learning have been rather scarce. Recent work has suggested to utilize the link between MARL and replicator dynamics from  population biology. Those equations have demonstrated very rich and complex behavior, such as sensitivity to initial conditions, and even   Hamiltonian chaos~\cite{Sato2002,Sato2003}.  A similar approach was used in~\cite{Tuyls2005}, where the Cross Learning model of~\cite{Borgers1997} was extended to stateless $Q$--learning, and where the link between multi--agent replicator dynamics and Evolutionary Game Theory was reiterated.  The replicator dynamics framework was also used in~\cite{Panait2008} to study advantages of lenient learners in repeated coordination games, where some convergence guarantees on certain games were obtained.  Furthermore, an attempt to generalize the stateless learning to a multi--state scenario was proposed in~\cite{Vrancx2008,Hennes2009}.

In addition to discrete strategy spaces, recent work has addressed games with continuous strategy spaces. For instance,  continuous strategy version of the prisoner dilemma has been considered in~\cite{Killingback1999,Wahl1999a,Wahl1999b,Le2007}. Replicator equations in continuous strategy spaces have been studied in the context of evolutionary and dynamical stability  in~\cite{Borgers1997,Oechssler2001,Cressman2005}. The corresponding replicator equation is similar to the one studied in the present paper, except there is no entropy (mutation) term. Most existing work in continuous strategy replicator dynamics has focused on steady state solutions that correspond to  Dirac's $\delta$--measures (monomorphic population), and in particular, their stability properties.  In contrast, here we will examine steady state solutions that have continuous support that exist at any non--zero exploration rate.

The work that is closest to one presented here is~\cite{Ruijgrok2005}, which studied continuous strategy replicator system with and without mutation. They established that for specific games, mutations resulted in non--trivial modification to the equilibrium structure (see also~\cite{Tuyls2009} for an extension of the approach to anisotropic  mutation mechanisms). The difference between the work above and the model presented here is the origin of the mutation term -- while in~\cite{Ruijgrok2005} the mutation was added  as a generic diffusive term,  in the model studied here it has a very intuitive entropic meaning that results from the Boltzmann action selection mechanism (see~\cite{Sato2003,Tuyls2003}). Specifically, the replicator equation in our case results from minimizing  a certain functional that is reminiscent of the so called {\em free energy} from statistical physics. This fact provides a very intuitive picture of the steady state structure.

\section{Model}
\label{sec:model}
We consider a system of $N$ agents that are playing repetitive  games with each other. In the present paper, we assume a stateless model, so that the reward for each agent depends only on the collective action of the agents. Let $x_i$ denote the action taken by the $i$-th agent. Also, let $x_{-i}$ denote the collective action profile of all the agents except $i$.  Without a loss of generality, we assume that the actions are restricted to the unit interval, $x_i \in [0,1],  i=1,2,..N$.

The game proceeds as follows: At each time step, each agent chooses an action, receives a reward that depends on the collective action of all the agents, and update his strategies accordingly.  Each agent has a $Q$--function that encodes the relative utility of  actions.  Those  Q--functions are updated after each time an agent selects an action, according to the following reinforcement rule~\cite{Watkins92}:
\begin{equation}
Q_i(x_i, t+\delta t) = Q_i(x_i, t) + \alpha \delta t [ f(x_i; x_{-i}) - Q_i(x_i, t)] 
\label{eq:q1}
\end{equation}
where $f_i(x_i; x_{-i}) $ is the reward of the agent $i$ when he takes the action $x_i$ and the rest of the agents take the collective action $x_{-i}$, and $\alpha$ is the learning rate.    

Next, we have to specify how agents choose actions based on $Q$--functions. Among several action--selection mechanisms, here we focus on the so called {\em Boltzmann exploration}~\cite{Kaelbling,SuttonBarto},   where the probability density of selecting a particular action $x_i$ is given as  
\begin{equation}
p_{i}(x_i, t) =  C_i(t) e^{\beta Q_i(x_i,t)} 
\label{eq:px}
\end{equation}
where $C_i(t)$ is simply a (time--dependent) normalization constant: 
\begin{equation}
  C_i(t) = \biggl [ \int dx_i e^{\beta Q_i(x_i,t)} \biggr ]^{-1}, 
\label{eq:cnorm}
\end{equation}
and where  $\beta\equiv 1/T$ is a parameter that controls the exploration/exploitation tradeoff, and has a meaning of inverse temperature owing to the analogy with statistical--mechanical systems. Note that in the zero temperature limit  $\beta\rightarrow \infty$, the distribution Equation~\ref{eq:px} becomes strongly peaked over those strategies which maximize the Q--function. This corresponds to the greedy selection mechanism, where agents do not perform any explorations over currently sub--optimal strategies. Conversely,  in the opposite (high temperature) limit $\beta \rightarrow 0$, the strategy profile becomes independent on the Q--function, and  strategies are chosen randomly with equal probability.

We now assume that the agents interact many times between two consecutive updates of their strategies. In this case, the reward of the $i$--th agent in Equation~\ref{eq:q1} should be understood in terms of the {\em average reward}, where the averaging is done over the strategies of other agents in the system. Specifically,   taking the limit $\delta t \rightarrow 0$, we can rewrite Equation~\ref{eq:q1} as follows: 
\begin{equation}
\frac{\partial Q_i(x_i, t)}{\partial t} =  \alpha [ r_i(x_i, t) - Q_i(x_i, t)]
\label{eq:q2}
\end{equation}
where $r_i(x_i,t)$ is the average reward ``felt" by the $i$--th agent:
\begin{equation}
r_i(x_i, t) = \int \dots \int  \prod_{j \neq i} dx_j p_j(x_j,t)f_i(x_i, x_{-i}) 
\label{eq:Rix}
\end{equation}

We want to eliminate $Q_i$--s from the dynamics, so that the learning trajectories are expressed solely through the agent strategies. To achieve this, let us take the derivative of Equation~\ref{eq:px} in respect to time:
\begin{equation}
\frac{\partial p_{i}}{\partial t} =\frac{\partial C_i}{\partial t}e^{\beta Q_i} + C_i\beta e^{\beta Q_i} \frac{\partial Q_i}{\partial t}\label{eq:px2}
\end{equation}
Note that 
\begin{equation}
\frac{dC_i}{d t} = - \frac{\int dx_i \beta e^{\beta Q_i} {\partial Q_i}/{\partial t}}{[ \int dx_i e^{\beta Q_i} ] ^2} \equiv -\beta C_i(t) \int dx _i p_i(x_i,t) {\partial Q_i}/{\partial t}
\label{eq:dcdt}
\end{equation}
Combining Equations~\ref{eq:px},~\ref{eq:q2}, and \ref{eq:dcdt}, we arrive at the following :
\begin{equation}
\frac{1}{p_i(x_i, t)} \frac{\partial p_i(x_i, t)}{\partial (\alpha \beta t)}  =  [ r_i(x_i, t) - {\bar r_i}(t) ]  - T [ \ln p_i(x_i,t) +  {\bar s_i}(t) ]
\label{eq:rep}
\end{equation}
where ${\bar r_i}(t) $ is the expected reward of agent $i$ averaged over the strategy profile $p_i(x_i,t)$, and ${\bar s_i}(t)$ is the entropy of that strategy profile:
\begin{equation}
{\bar r_i}(t) = \int d x_i r_i(x_i, t)p_i(x_i,t)  , \ , \  \bar{s_i}(t)=  \int d x_i p_i(x_i,t) \ln (p_i(x_i,t)) 
\label{eq:energyentropy}
\end{equation}


Equations~\ref{eq:rep} describe the mutual adaptation of interacting, Q--learning agents,  and allow a very simple interpretation. Indeed, the first term indicates that a probability of a playing a particular pure strategy increases with a rate proportional to the overall efficiency of that strategy. This is reminiscent of of fitness-based selection mechanism in population biology. The second, entropic term, on the other hand, does not have a direct analogue in population biology~\cite{Sato2003,Tuyls2003}. This term is due to the Boltzmann selection mechanism, and thus, describes the agents' tendency to explore sub--optimal (in the sense of their Q--function) strategies. Note that for $T=0$  the entropic term disappears and the system reduces to the conventional continuous--strategy replicator dynamics~\cite{Oechssler2001}. With a slight misuse of terminology, we will refer to Equations~\ref{eq:rep} as {\em replicator dynamics} even for $T>0$, when the entropic term is present.

\subsection{Steady State Solution}
\label{sec:steadystate}
We are interested in the asymptotic behavior of the Equations~\ref{eq:rep}  after sufficiently long time, $P_i(x_i)=p_i(x_i, t \rightarrow \infty)$. The steady--state equation is obtained  by setting the time--derivative to zero, which yields a set of equations (for each agent)
\begin{equation}
P_i(x_i) [ R_i(x_i)  - T \ln P_i(x_i)  +const] = 0, \ i=1,\cdots,N \ .
\label{eq:ss1}
\end{equation}
where $R_i(x_i)\equiv r_i(x_i, t\rightarrow \infty)$ depend on the collective strategy profile of all the agents as described by Equations~\ref{eq:Rix}, with $p_i(x_i, t)$ 
replaced by $P_i(x_i)$. Let us focus on solutions $P_i(x)$ that have continuous support on the interval $[0,1]$, i.e., $P_i(x)$ is strictly positive for all $x\in[0,1]$, except 
maybe a finite number of points. Since Equations~\ref{eq:ss1} should be satisfied for all the values $x_i\in[0,1]$, then the expression in the parenthesis should nullify for 
all $x$, which yields for $i=1,2,..N:$
\begin{equation}
P_i(x_i)  = A_i e^{\beta R_i(x_i) },
 \label{eq:ss2}
\end{equation}
where $A_i$-s are  constants to ensure proper normalization of strategies.

Equations~\ref{eq:ss2}  are coupled, highly non--linear integral equations, whose solution determine the steady state strategy profiles of the agents. The coupling enters non-trivially through the average rewards received by the agents.  Note, that in the single--agent learning scenario, $R_i(x)$ can be viewed as the reward received by the agent for playing the strategy $x$. In this case,  the steady state strategy profile given by Equation~\ref{eq:ss2}  can be interpreted as a Gibbs distribution for a statistical physical system with energy $-R_i(x)$ and temperature $1/\beta$, and can be derived from the following considerations. Let us define the following functional:
\begin{equation}
\Phi[p(x)] = -\int dx R(x)p(x) - T\int dxp(x)\ln(1/p(x))
\label{eq:functional}
\end{equation}
$\Phi$ is closely related to  the concept of {\em free energy} from equilibrium statistical physics. It is easy to check that Equation~\ref{eq:ss2} minimizes $\Phi$, subject to the condition that $p(x)$ is normalized. Indeed,  introducing a Lagrange multiplier to account for the normalization constraint, and taking the functional derivative with respect to $p$, we obtain
\begin{equation}
\delta \Phi[p(x)] = -\int dx [R(x) - T \ln p(x)) - const]\delta p(x) =0
\label{eq:functional_var}
\end{equation}
which again yields~\ref{eq:ss2}. This suggests that at non-zero temperature, the replicator dynamics will lead to a steady state that has non-zero entropy. Thus, generally speaking, the steady state solution of the replicator system will not correspond to any Nash equilibria, if those  equilibria are monomorphic. In the terminology of population biology, non--zero temperature guarantees population diversity, and the monomorphic solutions are not allowed. Note that the actual degree of diversity if governed by the temperature.  We expect that by decreasing the temperature, the steady state will converge towards certain Nash equilibrium of the game.  Below we examine this question in more detail.

\section{Examples}
\label{sec:example}
Owing to the highly non--linear nature of the steady state equations, they cannot be solved analytically in the most general case of arbitrary payoffs. However,  one can still establish important results for certain  class of games. In the reminder of the paper we focus on several such examples. We will limit our consideration to two--player games. Let $x$ and $y$ denote the actions of each agent. Then the steady state equations have the following form:  
\begin{eqnarray}
P_1(x)  &=& A_1 e^{\beta R_1(x) } \equiv A_1 e^{\beta \int dy f_X(x,y)P_2(y) }\\
P_2(y)  &=& A_2 e^{\beta R_2(y) } \equiv A_2 e^{\beta \int dx f_Y(x,y)P_1(x) }
 \label{eq:sstwoplayer}
\end{eqnarray}
where $A_1$, $A_2$ are normalization constants. 
\comment{
In case when the system allows a symmetric solution, one has
\begin{equation}
P(x)  = A_1 \frac{ e^{\beta \int dy f(x,y)P(y) } } {  \int dx e^{\beta \int dy f(x,y)P(y) }  }
 \label{eq:sol}
\end{equation}
}

Below we will examine the above system for a number of games. We will compare the steady state analysis of the replicator equations with the equilibrium  strategies of Q--learning agents obtained through Monte--Carlo simulations. The latter was generated by having two agents interacting with each other through the corresponding game dynamical payoff matrix, and adapting their strategies according to the Q--learning mechanism Equations.~\ref{eq:q2} and~\ref{eq:px} until they reach an equilibrium. We observed that the simulations results  generally agreed well with the replicator model, provided that the simulations parameters such as learning rate and strategy discretization were chosen carefully. Generally, we found that the larger the $\beta$, the smaller the learning rate should be in order to reproduce the behavior predicted by the replicator analysis. 

\subsection{Bi--Linear Payoff}
\label{sec:bilinear}
We first consider games with symmetric bi--linear payoffs structure that has the following form:
\begin{equation}
f_X(x,y)=axy + bx \ , \  f_Y(x,y)=axy + by ,\ a \neq 0 \ .
\end{equation}
This game has a number of different pure Nash equilibria depending on the parameters $a$ and $b$. For instance, if $a>0$,  then the game has a (symmetric)  pure Nash equilibrium at  $(x,y)=(1,1)$ or $(x,y)=(0,0)$, depending on whether $b\ge -a$ or $b < -a$. For $a<0,0<b<|a|$, there are two asymmetric NE at $(0,1)$ and $(1,0)$. Finally, for $ab<0, |b|<|a|$ there is a pure Nash equilibrium at $x=y=|b|/|a|$, and furthermore, any strategy profile with a mean  $|b|/|a|$ is a Nash equilibrium as well.

To find the steady state strategy profiles, we note that the average reward of the agents $x$ (y) is a linear function of $x$ (y):
\begin{eqnarray}
R_1(x) =x(a{\overline y} + b) \ , \ R_2(y) =y(a{\overline x} + b) 
\end{eqnarray} 
where ${\overline x},{\overline y} $ are the means of respective strategy profiles.  Then, using Equation~\ref{eq:ss2}, we find that the steady state solution should have the following exponential form:  
\begin{equation}
P_1(x)=\frac{\gamma_1}{e^{\gamma_1}-1} e^{\gamma_1 x}, P_2(y)=\frac{\gamma_2}{e^{\gamma_2}-1} e^{\gamma_2 y} 
\label{eq:linear_steady}
\end{equation}
where we have defined
\begin{equation} 
\gamma_1 =\beta (a{\overline y} + b) \ , \  \gamma_2 =\beta (a{\overline x} + b).
\label{eq:gamma12}
\end{equation}  
Next, we calculate  the means of the steady state strategy profiles
\begin{equation}
{\overline x} \equiv \int dx x P_1(x) = \frac{1}{1-e^{-\gamma_1}}-\frac{1}{\gamma_1} \ , \ {\overline y} \equiv \int dy y P_2(y) = \frac{1}{1-e^{-\gamma_2}}-\frac{1}{\gamma_2}
\label{eq:means}
\end{equation}
Finally, inserting Equations~\ref{eq:means} into  Equation~\ref{eq:gamma12} to arrive at the following self--consistency equations for $\gamma_1$, $\gamma_2$:
\begin{equation}
\frac{\gamma_1} { \beta} = g(\gamma_2) \ , \  \frac{\gamma_2} { \beta} = g(\gamma_1)
\label{eq:selfconsist}
\end{equation}
where the fuction $g(\gamma)$ is defined as follows:
\begin{equation}
g(\gamma) = b + a\biggl [\frac{1}{1-e^{-\gamma}}-\frac{1}{\gamma}\biggr ]
\label{eq:g}
\end{equation}
Thus, the steady state strategy profiles are given by Equations~\ref{eq:linear_steady}, where $\gamma_1, \gamma_2$ need to be determined by solving the system of transcendental  equations~\ref{eq:selfconsist}. As we show below, the nature of the solution strongly depends on the choice of the parameters. We now examine this dependence in more detail.  

First, let us focus on the symmetric solutions, $\gamma_1 = \gamma_2\equiv \gamma$, in which case Equation~\ref{eq:selfconsist} becomes
\begin{equation}
\frac{\gamma}{\beta} = g(\gamma)
\label{eq:symm}
\end{equation}

\begin{figure}[!h]
\centering
\subfigure[]{
    \includegraphics [width=0.35\textwidth] {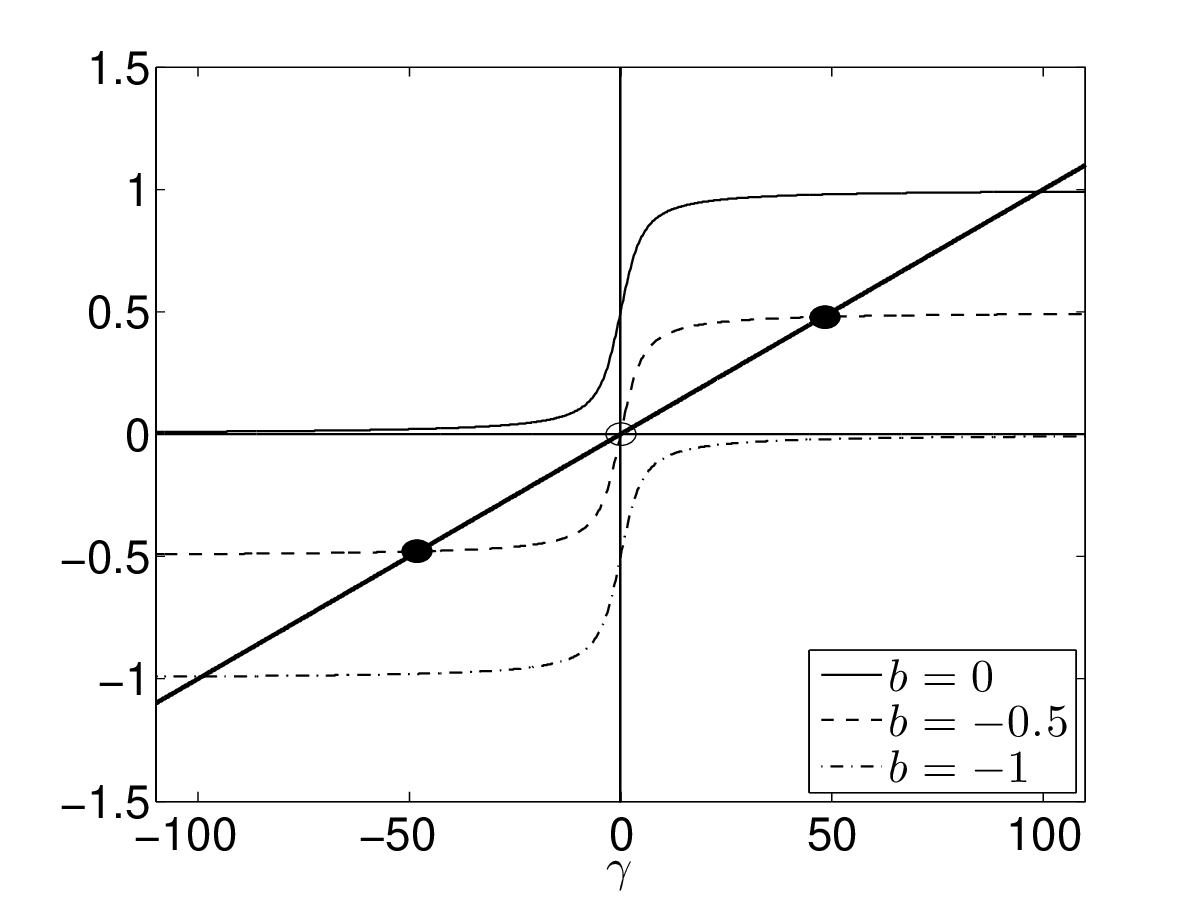}\label{fig:Lin1a}
    }
    \subfigure[]{
    \includegraphics[width=0.35\textwidth]{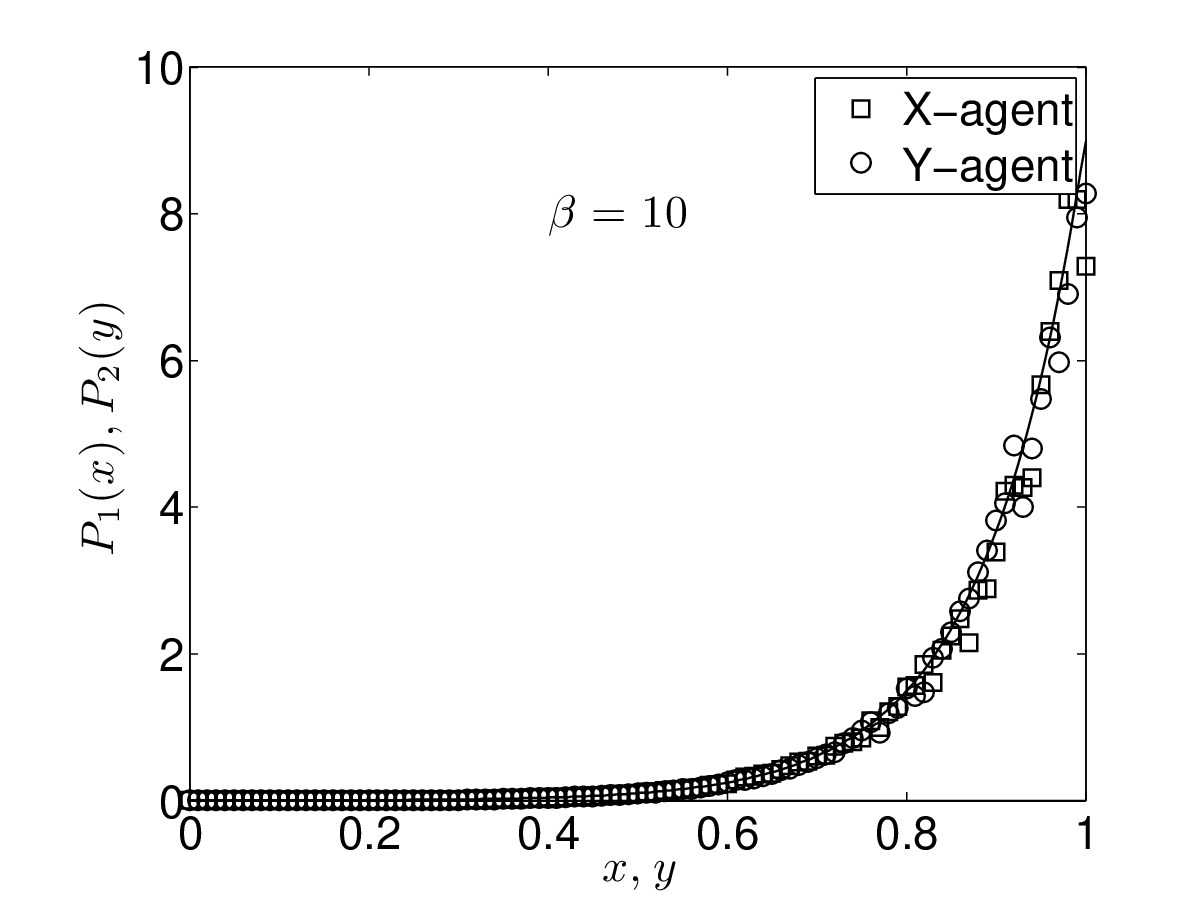} \label{fig:Lin1b}
    }
     \caption{(a) Graphical representation of Equation~\ref{eq:symm} for  $a=1$  and different values of $b$, as shown. The straight line has a slope $\frac{1}{\beta}$; (b) Steady state strategy profiles for  $a=1$, $b=0$. The solid line is the analytical result. The symbols represent the result of simulations. }
     \label{fig:Lin1}
\end{figure}

Graphical illustration of Equation~\ref{eq:symm} is presented in Figure~\ref{fig:Lin1}, where we plot $g(\gamma)$ together with the  line $\gamma/\beta$ for different values of $b$.  Let us first consider the case $a>0$ (Figure~\ref{fig:Lin1a}). In this case,  $g(\gamma)$ is a monotonically increasing function of its argument, which tends to $b$ as $\gamma \rightarrow -\infty$, and to $b+a$ as  $\gamma \rightarrow \infty$. A simple inspection shows that for $b>0$, the equation has a unique solution that increases almost linearly with the inverse temperature $\beta$. Similarly, for  $b\le-a$, the equation again has  a unique solution, which decreases linearly with increasing $\beta$. Thus, the steady state strategies are peaked around $x,y=0$ ($x,y=1$) for $b<-a$, ($b>0$), with the width of the distribution shrinking linearly with $\beta$. 
\begin{figure}[!h]
\centering
    \subfigure[]{
    \includegraphics[width=0.35\textwidth]{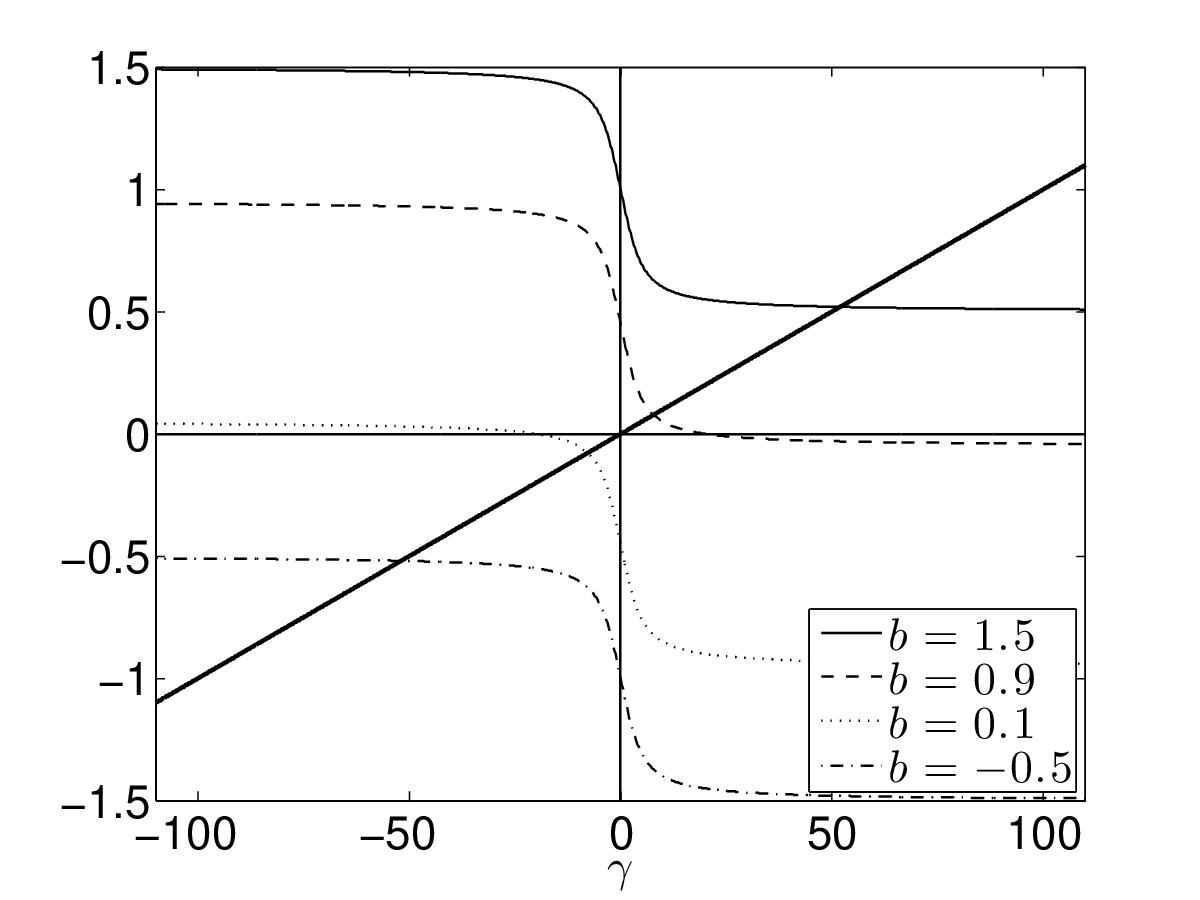} \label{fig:Lin2a}
    }
     \subfigure[]{
    \includegraphics[width=0.35\textwidth]{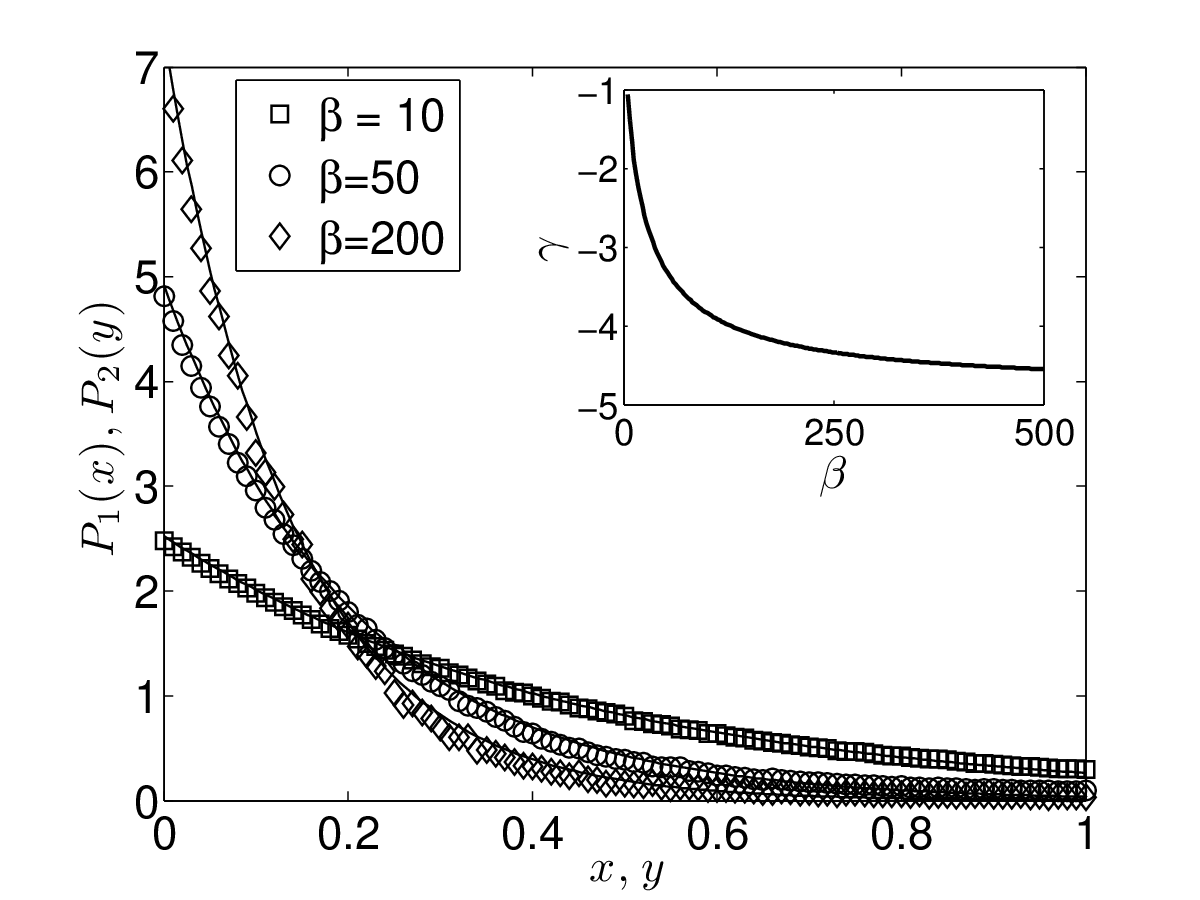} \label{fig:Lin2b}
    }     \caption{(a) Graphical representation of Equation~\ref{eq:symm} for  $a=-1$  and different values of $b$, as shown; (b) Steady state profile for $a=-1$, $b=0.1$, and different values of the inverse temperature $\beta$. The solid line is the analytical result. The symbols represent the result of simulations. }
     \label{fig:Lin2}
\end{figure}

The situation is different for  the intermediate values of $b$. Indeed, it can be seen from Figure~\ref{fig:Lin1a}, that for  $-a<b<0$, there is a critical value $\beta_c$ so that for any $\beta>\beta_c$, there are three distinct solutions. Thus, the replicator equation has three different steady state solutions given by Equations~\ref{eq:linear_steady}. Our simulations indicate that  the steady state profile corresponding to the solution in the middle (open circle in Figure~\ref{fig:Lin1a}) is unstable; see the appendix where  the non-stability  is illustrated analytically for a specific choice of $a$ and $b$. Our simulations also suggest that  the other two solutions (filled circles in Figure~\ref{fig:Lin1a}) are stable, and one arrives at either of those solutions depending on the initial conditions. Furthermore, it is easy to check that asymptotic behavior of those solutions for large values of $\beta$ is $\gamma \approx \pm \beta  |b|$.  Thus, in zero--temperature limit $\beta \rightarrow \infty$ the steady state strategy profiles are $\delta$--functions at either $0$ or $1$, depending on the initial conditions.  This prediction was confirmed in our simulations in Figure~\ref{fig:Lin1b}, where we show the steady--state strategy profile for a game with $a=1$ and $b = 0$. The symbols are the results of Monte--Carlo simulations, which agree very well with the analytical prediction. For this particular case, the steady--state strategy is peaked around the Nash equilibrium point $x=y=1$. Following the discussion above, it is clear that while increasing $\beta$, $p(x)$ will tend to a point mass at $x=1$.

We also note that asymptotic behavior of the solutions for large values of $\beta$ is $\gamma \approx \pm \beta  |b|$.  Thus, in zero--temperature limit $\beta \rightarrow \infty$ the steady state strategy profiles are $\delta$--functions at either $0$ or $1$, depending on the initial conditions.  This prediction was confirmed in our simulations in Figure~\ref{fig:Lin1b}, where we show the steady--state strategy profile for a game with $a=1$ and $b = 0$. The symbols are the results of Monte--Carlo simulations, which agree very well with the analytical prediction. For this particular case, the steady--state strategy is peaked around the Nash equilibrium point $x=y=1$. Following the discussion above, it is clear that while increasing $\beta$, $p(x)$ will tend to a point mass at $x=1$.

Now consider the case $a<0$. The corresponding $g(\gamma)$ is shown in  Figure~\ref{fig:Lin2a}. Since $g(\gamma)$ is a strictly decreasing function, Equation~\ref{eq:symm} has only one solution, for any values of  $b$ and $\beta$. However, a simple analysis reveals that depending on the value of $b$, the asymptotic behavior of the solution for large $\beta$ is different.  Indeed, for $|b|>|a|$, the  solution behaves asymptotically as $\gamma \approx \pm |b|\beta$. Thus,  the zero--temperature  steady state strategy profile again corresponds to point mass at $x=0$ or $x=1$, depending on the initial conditions. For the intermediate values of $|b|<-a$, on the other hand, the solution is almost independent of $\beta$, as it can be seen from the graph. In this case, the steady state strategies have continuous support for any $\beta$. This behavior is depicted in Figure~\ref{fig:Lin2b}, where we plot the steady state strategy profile  for different values of $\beta$. Note that increasing $\beta$ makes the distribution more peaked initially, but it saturates for sufficiently large $\beta$. This is demonstrated in the inset of Figure~\ref{fig:Lin2b}  which shows the dependence of $\gamma$ on the inverse temperature. Thus, even in the limit $\beta \rightarrow \infty$, the solution will always have a finite (non--zero) entropy. Furthermore, we observe that the respective means of the strategy profiles converges towards ${\overline x} ={\overline y} =|b|/|a|$ as $\beta$ increases. Thus, the steady state of the replicator dynamics corresponds to the mixed Nash equilibria of the game as temperature goes to zero.

\begin{figure}[!h]
\centering
     \subfigure[]{
    \includegraphics[width=0.32\textwidth]{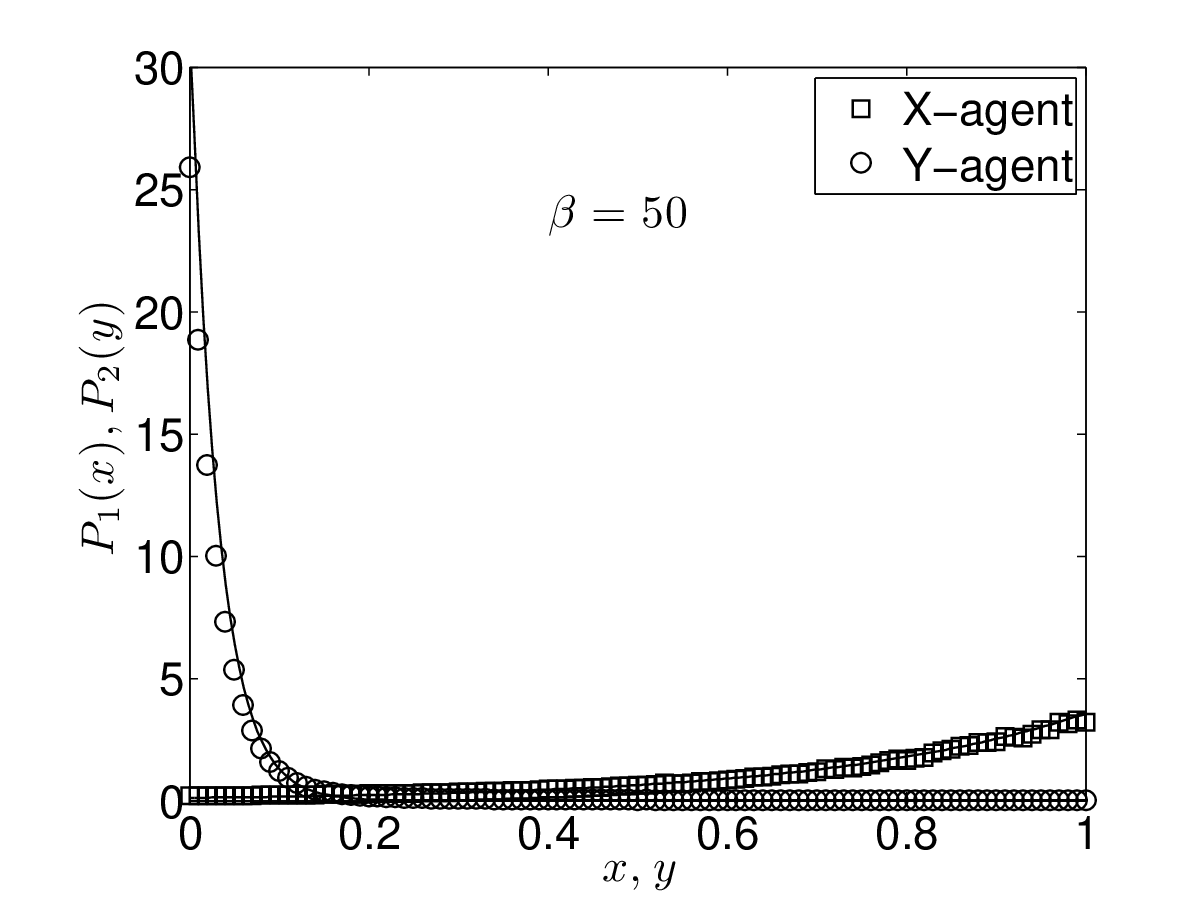} \label{fig:Lin3a}
    }
     \subfigure[]{
    \includegraphics[width=0.32\textwidth]{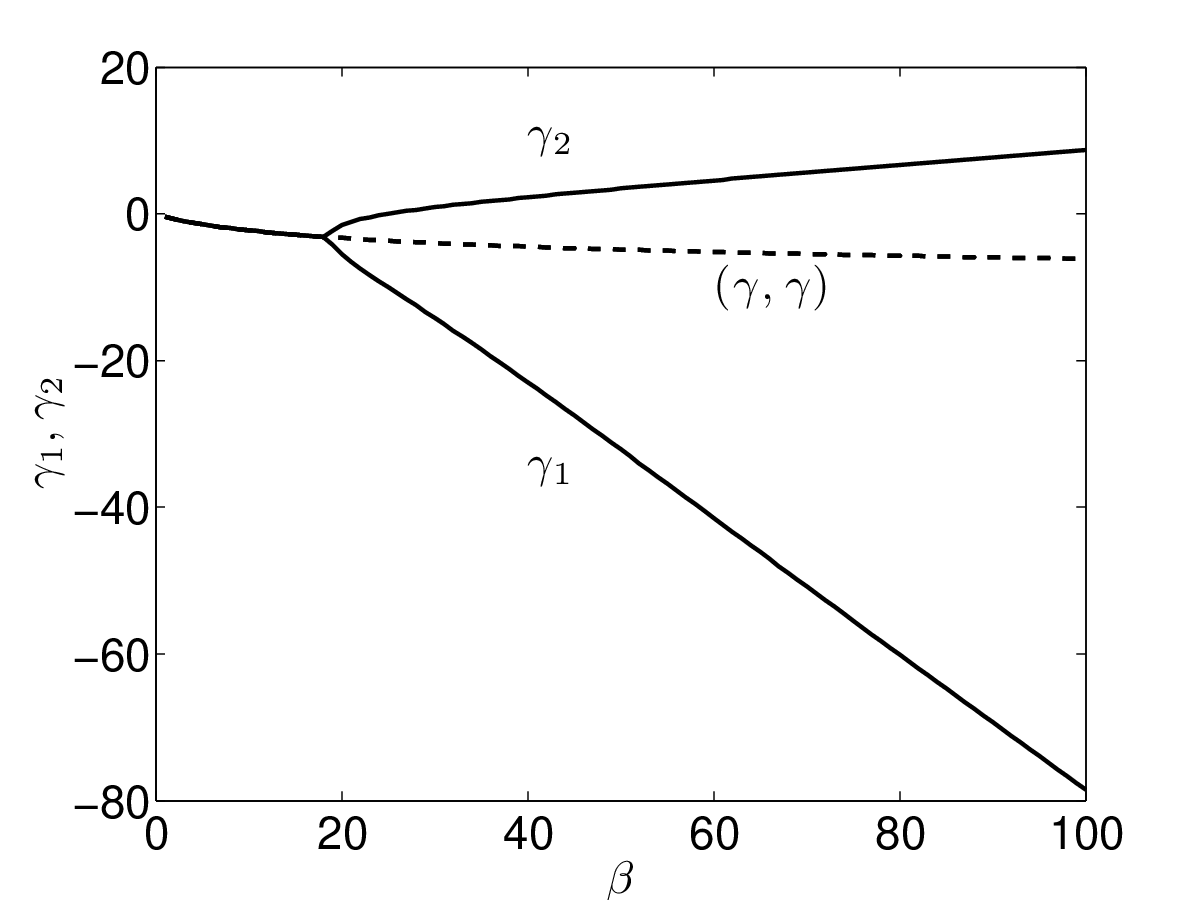} \label{fig:Lin3b}
    }
     \subfigure[]{
     \includegraphics[width=0.32\textwidth]{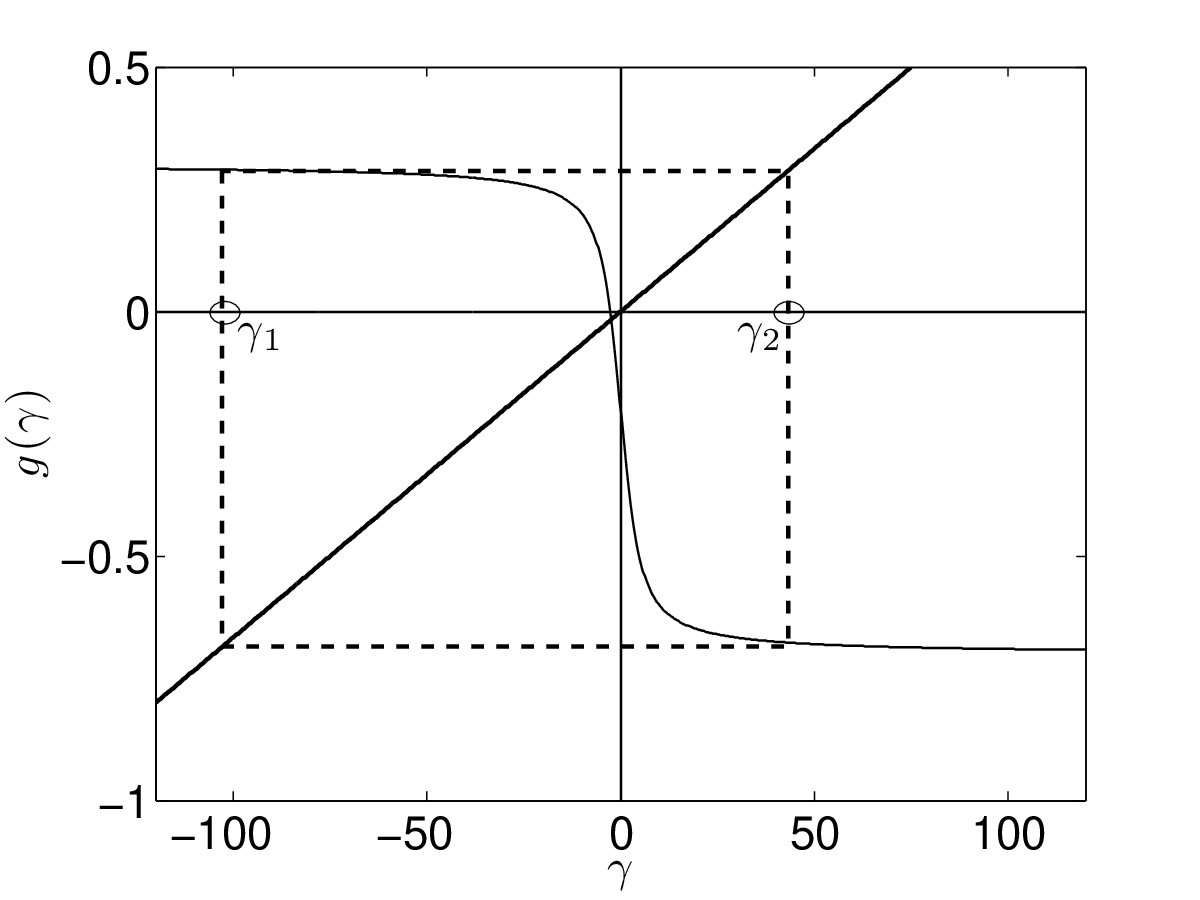}  \label{fig:Lin4}
     }
     \caption{(a) Asymmetric steady--state strategy profiles for the bi--linear game. The solid line is the analytical result, and the symbols represent the simulation results; (b) The ``bifurcation" diagram of the equation~\ref{eq:selfconsist}; (c) Illustration of a cycle with period $l=2$. The circle represent corresponding solutions $\gamma_1$, $\gamma_2$.  }
     \label{fig:Lin3}
\end{figure}

So far our analysis has focused on the symmetric solutions to the steady state equations~\ref{eq:selfconsist}.  For sufficiently small $\beta$, the symmetric solution is the only one. However, above some critical value of $\beta$, another, asymmetric solution   appears (strictly speaking, there are two solutions related by $x\leftrightarrow y$ symmetry). This is shown in Figure~\ref{fig:Lin3a}, where we compare analytical and simulation results for a  particular value of $\beta$. In simulations, we had to start from asymmetric initial strategies in order to arrive at the asymmetric steady state. The structure of the solution can be summarized as follows: While increasing $\beta$, the strategy  around $0$ narrows much faster compared to the strategy around $1$. This is also shown in Figure~\ref{fig:Lin3b}, where we plot the bifurcation diagram of the solution to Equations~\ref{eq:selfconsist}. For small $\beta$, there is only a symmetric solution, $\gamma_1=\gamma_2\equiv \gamma$. Starting from a critical $\beta$, an asymmetric solution appears. In the limit $\beta \rightarrow \infty$, one has $\gamma_1 \rightarrow -\infty$, $\gamma_2 \rightarrow \infty$, so that corresponding strategy profiles Equations~\ref{eq:linear_steady} converge to one of the pure asymmetric NE. We also note that the symmetric solution exists at any value of $\beta$, as shown by the dashed line.

To understand the nature of the asymmetric solution, we make the following  observation: Consider the  discrete map 
\begin{equation}
z_{t+1} = \beta g(z_t)
\label{eq:map}
\end{equation}
Clearly, the symmetric solutions to Equation~\ref{eq:selfconsist} correspond to the fixed points of this map, $z^*=\beta g(z^*)$. Furthermore, it is easy to see that asymmetric solutions  of Equation~\ref{eq:selfconsist}, if they exist,  correspond to the cycles of  the map~\ref{eq:map} with period $l=2$. And conversely, if Equation~\ref{eq:selfconsist} allows an asymmetric solution, then the map~\ref{eq:map} necessarily has a cycle with period $l=2$. A simple inspection shows that  for $a<0$, $|b|<|a|$ such cycle exist for certain $\beta$, as shown schematically in Figure~\ref{fig:Lin4}.  Note that the asymmetric solution appears at some critical value of $\beta$. To understand why this is the case,  recall that the steady state profile is the minimizer of the free energy. When $\beta \rightarrow 0$, the minimization of the free energy is analogous to  maximizing the entropy, which results in a uniform distributions with $\gamma_1=\gamma_2 = 0$. For finite but sufficiently small $\beta$, the minimization is still dominated by the entropy term, thus, the steady state profiles for both agents will be close to the the uniform distribution.  If one increases $\beta$, however, the role of the entropic term decreases,  until, at some critical $\beta$  the second asymmetric solution appears.   
%

\subsection{Quadratic Payoff}
We now consider  another important class of games where the payoffs can be expressed through the following quadratic forms:
\begin{eqnarray}
f_X(x,y)&=&-(x+y-2a_1)^2\nonumber \\
f_Y(x,y)&=&-(x+y-2a_2)^2
\label{eq:quadraticpayoff}
\end{eqnarray}
Here $0<a_1,a_2 < 1$, and generally speaking, $a_1\neq a_2$. Let $P_1(x)$ and $P_2(y)$ be the steady state strategy profiles corresponding to this payoff. It is easy to check that the average rewards are given as follows:
\begin{eqnarray}
R_1(x)&=&-(x+\overline{y} -2a_1)^2  - (\overline {y^2}  -{\overline y}^2 ) \\
R_2(y)&=&-(y+\overline{x} -2a_2)^2 -  (\overline {x^2}  -{\overline x}^2 )
\label{eq:quadraticRXRY}
\end{eqnarray}
This suggests the following form for the steady state strategies:   
\begin{eqnarray}
P_1(x)&=&c(x_0) e^{-\beta(x-x_0)^2} \\
P_2(y)&=&c(y_0) e^{-\beta(y-y_0)^2}
\label{eq:quadratic_steady}
\end{eqnarray}
where $c(x_0)$, $c(y_0)$, are the respective normalization factors, and the function $c(z)$ can be expressed through the error functions as follows:
\begin{equation}
c(z)=2\sqrt{\frac{ \beta}{\pi}} [erf(\sqrt{\beta}z)+erf(\sqrt{\beta}(1-z))]^{-1}  
\label{eq:quadratic_norm}
\end{equation}
Combining Equations~\ref{eq:sstwoplayer},~\ref{eq:quadraticRXRY} and \ref{eq:quadratic_steady}, we find the following transcendental equations for the parameters $x_0$ and  $y_0$:
\begin{eqnarray}
x_0&=&2a_1 - \mu(y_0)
\label{eq:quadratic_norm1} \\
y_0&=&2a_2 - \mu(x_0) 
\label{eq:quadratic_norm2}
\end{eqnarray}
where the function $\mu(z)$ is the mean of a truncated Gaussian distribution centered at $z$, and is given as follows:
\begin{equation}
 \mu(z) =z - c(z)\frac{e^{-\beta(1-z)^2}-e^{-\beta z^2}}{2\beta}  
 \label{eq:quadratic_means} 
\end{equation}

Let us first analyze the symmetric case $a_1=a_2\equiv a$. Note that in this case the game becomes a pure coordination game, which has continuously many Nash equilibria given by $x^*+y^*=2a$. Furthermore, it is easy to check that there is no mixed Nash equilibrium. To see why this is the case, assume the contrary, and let $P_1(x)$ be the mixed NE strategy of the first agent. Then, using  Equation~\ref{eq:quadraticRXRY}, it is straightforward to show that the best response of the second agent  is to play a pure strategy $P_2(y)=\delta(y-y^*)$ with $y^* = 2a - \overline{x}$. However, if the second agent plays according to this pure strategy, then the first agent will do better by playing a pure strategy as well, $P_1(x) = \delta (x-\overline{x})$.

\begin{figure}[!h]
\centering
 \subfigure[]{
    \includegraphics[width=0.35\textwidth]{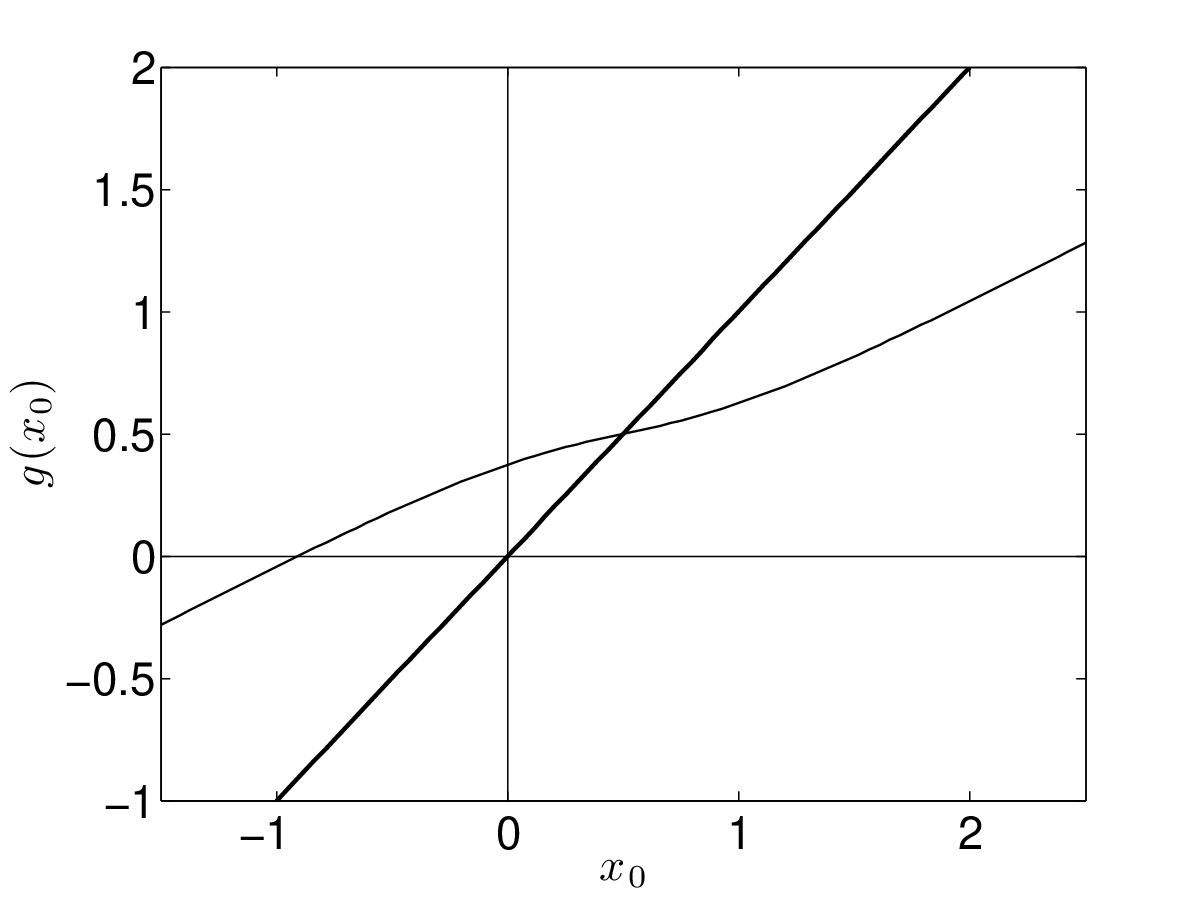} \label{fig:quad5a}
    }
     \subfigure[]{
    \includegraphics[width=0.35\textwidth]{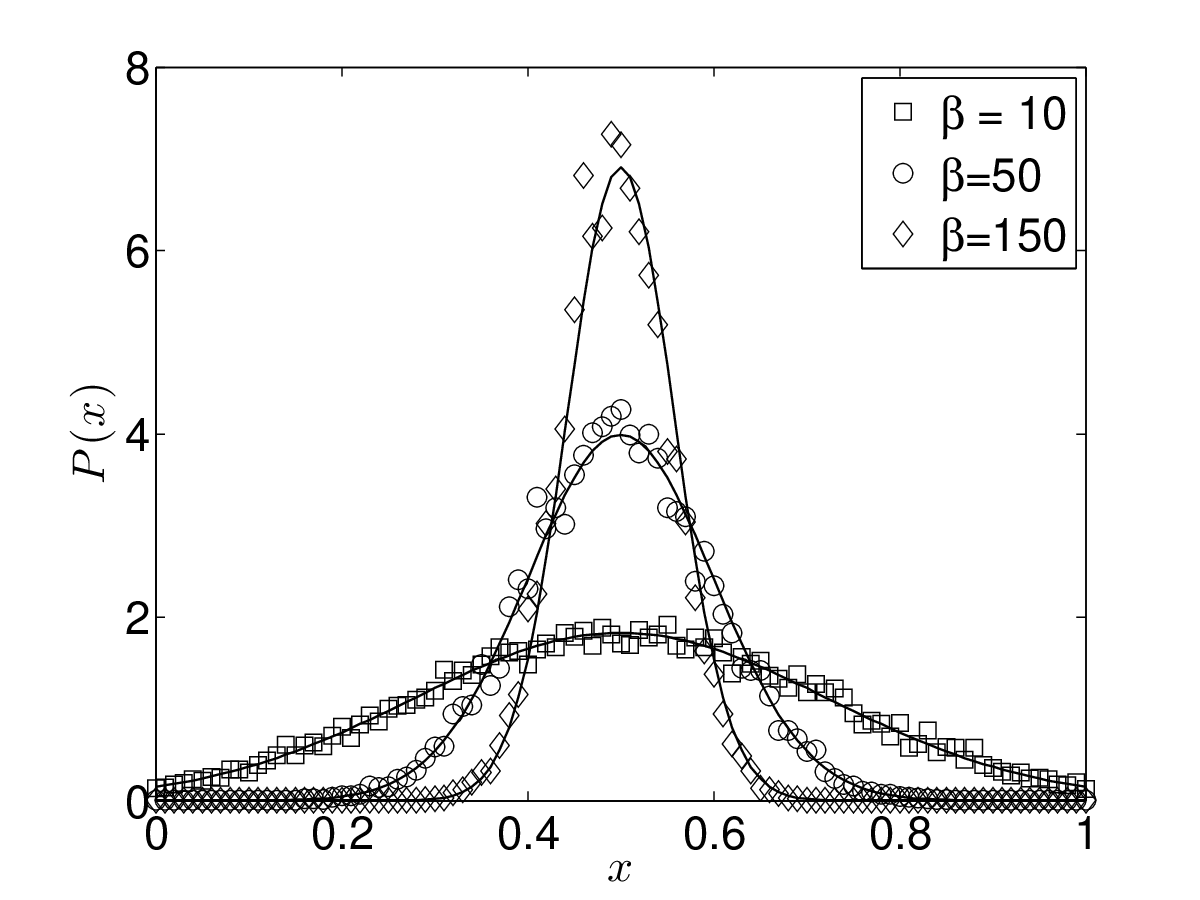} \label{fig:quad5b}
    }
     \caption{(a) Graphical representation of Equation~\ref{eq:x0_symm}; (b) Steady--state strategy profiles (analytical and simulations) for the coordination game with $a=0.5$ and for three different values of the inverse temperature $\beta$.}
     \label{fig:quad}
\end{figure}

Let us now consider the corresponding steady state structure within the replicator framework. The steady state strategy profiles are  (truncated) normal distributions centered at points $(x_0,y_0)$ which need  to be found from the system of transcendental Equations~\ref{eq:quadratic_norm1}--~\ref{eq:quadratic_means}. We now analyze this system in more detail. First of all, we check  for a symmetric solution $x_0=y_0$, for which the system of equations reduces to the following equation: 
\begin{equation}
x_0 = 2a - \mu(x_0)\equiv a + c(x_0)\frac{e^{-\beta(1-x_0)^2}-e^{-\beta x_0^2}}{4\beta}
\label{eq:x0_symm}
\end{equation}
Graphical representation of Equation~\ref{eq:x0_symm} is shown in Figure~\ref{fig:quad5a}. An inspection confirms that a symmetric solution is present for any value of the inverse temperature $\beta$. The actual steady state strategy profiles corresponding to different values of $\beta$ are shown in Figure~\ref{fig:quad5b}, together with simulation results.  Note that in the limit $\beta \rightarrow \infty$ one has $x_0=y_0\equiv a$, suggesting that the strategy profiles of both agents become peaked around $a$ in the limit of large $\beta$.

\begin{figure}[!h]
\centering
    \includegraphics [width=0.35\textwidth] {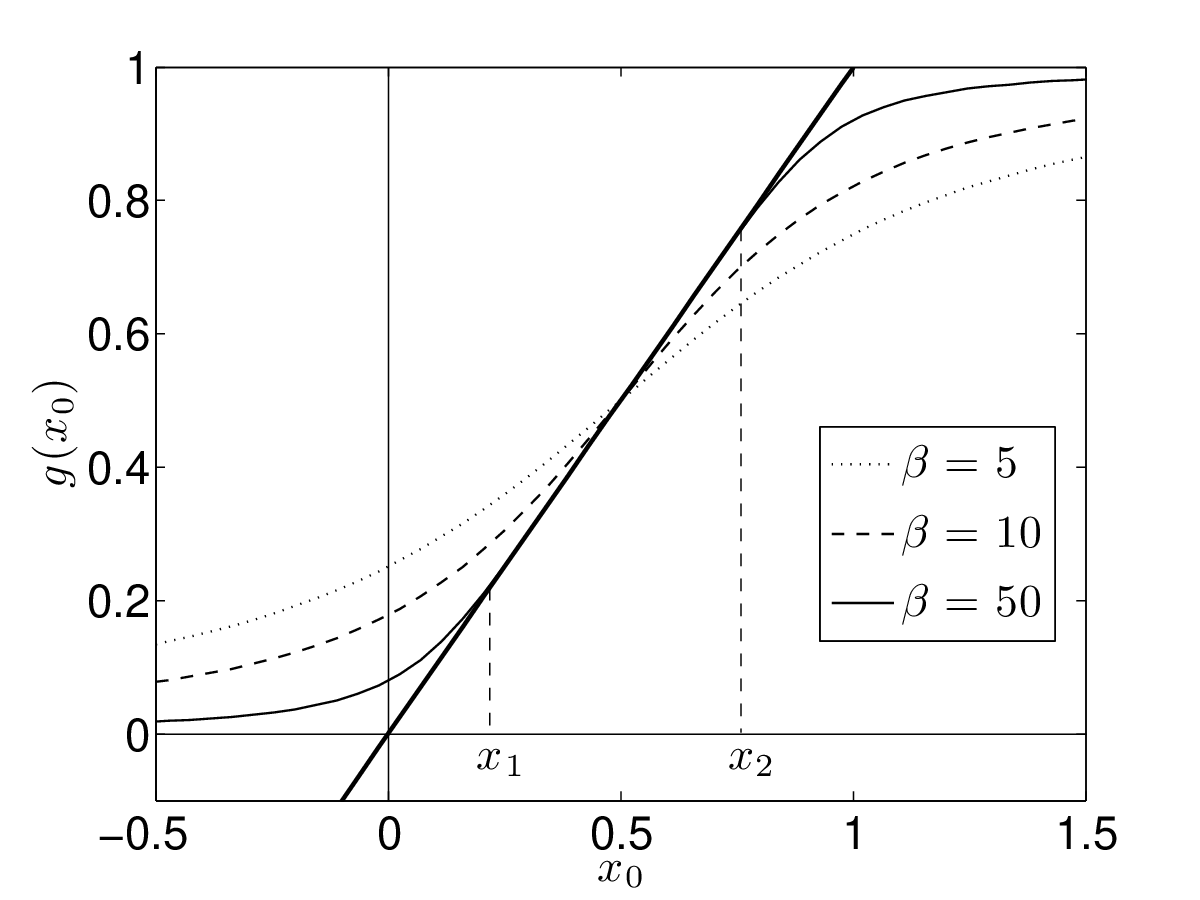}
     \caption{Graphical representation of Equation~\ref{eq:x0_nonsymm} shown for three different values of the parameter $\beta$.}
     \label{fig:quad6a}
\end{figure}

The picture above suggests that in the zero temperature limit $\beta \rightarrow \infty$, the steady state strategy profiles sholud correspond to the symmetric pure strategy Nash equilibrium $x^*=y^*=a$. Indeed, our results suggests that if one starts in the vicinity of the symmetric equilibrium, then this picture holds.  However, further analysis reveals that the convergence to this symmetric equilibrium is not trivial. To understand why this is the case, let us consider again the steady state equations~\ref{eq:quadratic_norm1} and~\ref{eq:quadratic_norm2}, and  look for the solutions of form $x_0+y_0=2a$. This leads to the following  equation: 
\begin{equation}
x_0 = 2a - \mu( 2a - x_0)
\label{eq:x0_nonsymm}
\end{equation}
The graphical representation of Equation~\ref{eq:x0_nonsymm}  is  shown in Figure~\ref{fig:quad6a}. An inspection of this equation for various $\beta$  yields the following observation: the symmetric solution $x_0=y_0=a$ is the only solution, and for relatively  small values of $\beta$,  the replicator dynamics settles into this symmetric solution relatively quickly. However, increasing $\beta$ leads to the appearance of continuously many {\em metastable } states that are characterized by the condition $x+y\approx 2a$.  Thus, when the system starts close to those metastable states, it can get trapped there for very long times,  converging very slowly  to the actual steady state. In fact, the convergence times diverges as $\beta \rightarrow \infty$.  Also, note that the metastable state that the system will fall into will depend on the initial conditions. 

\begin{figure}[!h]
\centering
    \subfigure[]{
    \includegraphics[width=0.35\textwidth]{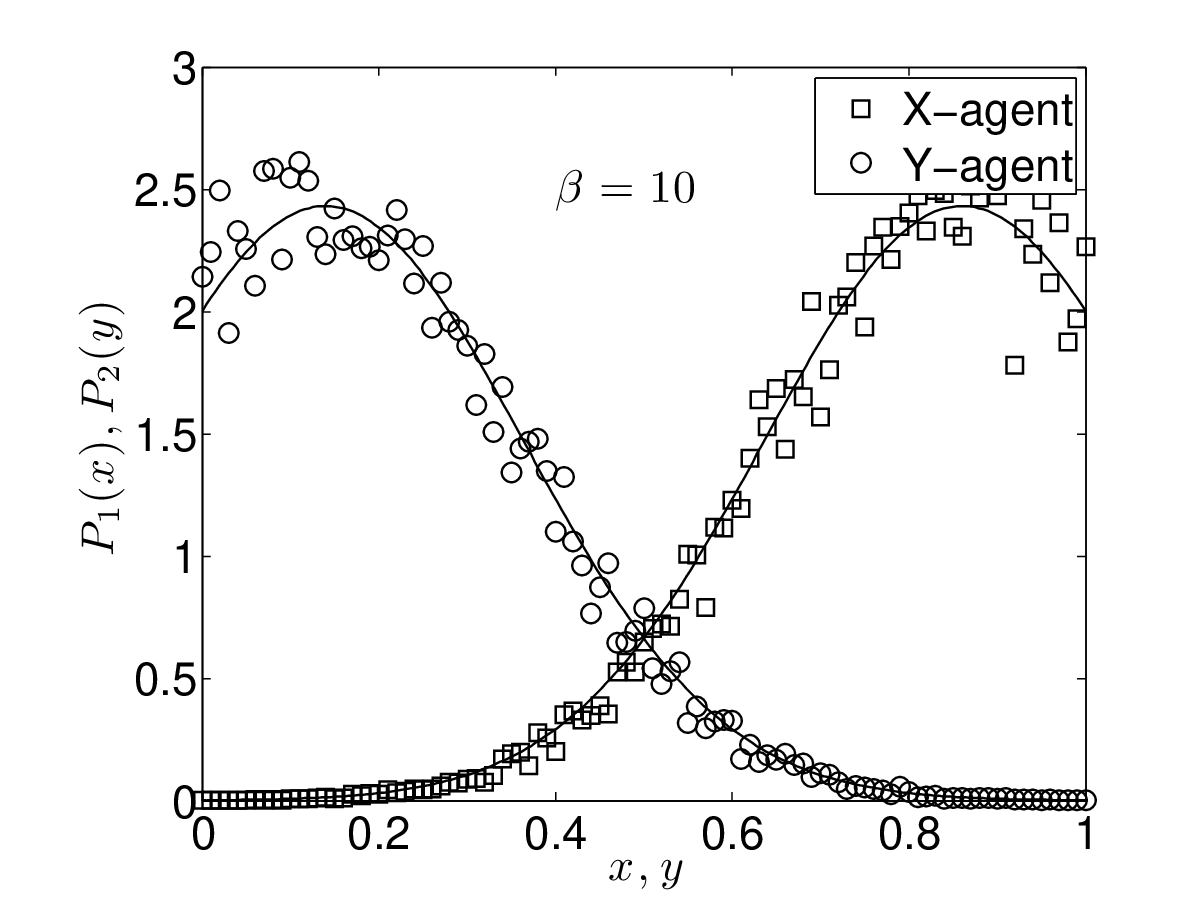} \label{fig:quad7a}
    }
     \subfigure[]{
    \includegraphics[width=0.35\textwidth]{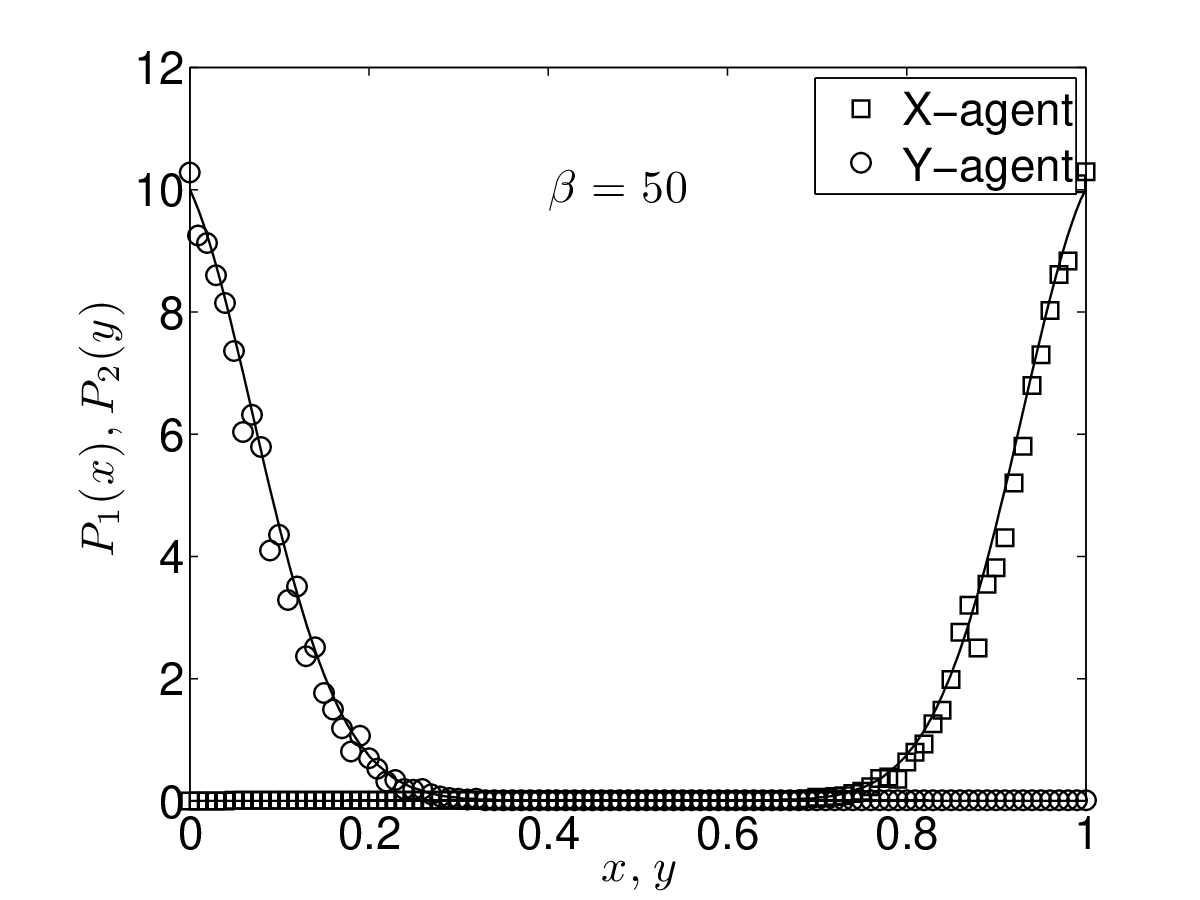} \label{fig:quad7b}
    }     \caption{Steady--state strategy profiles for the asymmetric quadratic  game  with parameters $a_1=0.45$, $a_2=0.55$, and for two values of the inverse temperature $\beta=10$ (a) and $\beta=50$ (b). }
     \label{fig:quad7}
\end{figure}

Finally, we consider the asymmetric case $a_1\neq a_2$. It can be shown that even small asymmetry can drastically change the structure of the Nash equilibria. Consider, for instance, the following perturbation of the game considered above: $a_1=a-\delta$, $a_2=a+\delta $, with  $a=0.5$, and where $\delta>0$ is a small positive constant. It is clear that for any positive $\delta $, no matter how small, the asymmetry leads to a single pure Nash equilibrium at $\{x^*,y^*\}=\{0,1\}$. Our simulations show that similar behavior persists for finite $\beta$, and that the agents' strategies drift towards the deterministic equilibrium points when one increases $\beta$, as shown in Figure~\ref{fig:quad7}.

\subsection{The Political Advertisement Game}
In addition to simple bi--linear and quadratic games, where analytical examination of the steady state structure was possible, we considered other games for which the solution of the steady state equation cannot be obtained analytically, so one has to use numerical techniques.  As an example, we consider the so called {\em political advertisement} game, were two political parties  decide their expenditure levels $x$ and $y$ for campaign advertisement. The total number of votes  participating in the election  is proportional to the collective expenditure, and the fraction of votes each party gets  is proportional to individual expenditures. Thus, the payoff  has the following structure:
\begin{equation}
f_X(x,y)  = \frac{x}{x+y} - x, ~~~ f_Y(x,y)  = \frac{y}{x+y} -y, 
 \label{eq:polad}
\end{equation}
It is easy to show that this game has a single non--trivial Nash equilibrium at $x^*=y^*=1/4$. Furthermore, a simple inspection  shows that there are no mixed Nash equilibria. 

\begin{figure}[!h]
\centering
\subfigure[]{
    \includegraphics [width=0.35\textwidth] {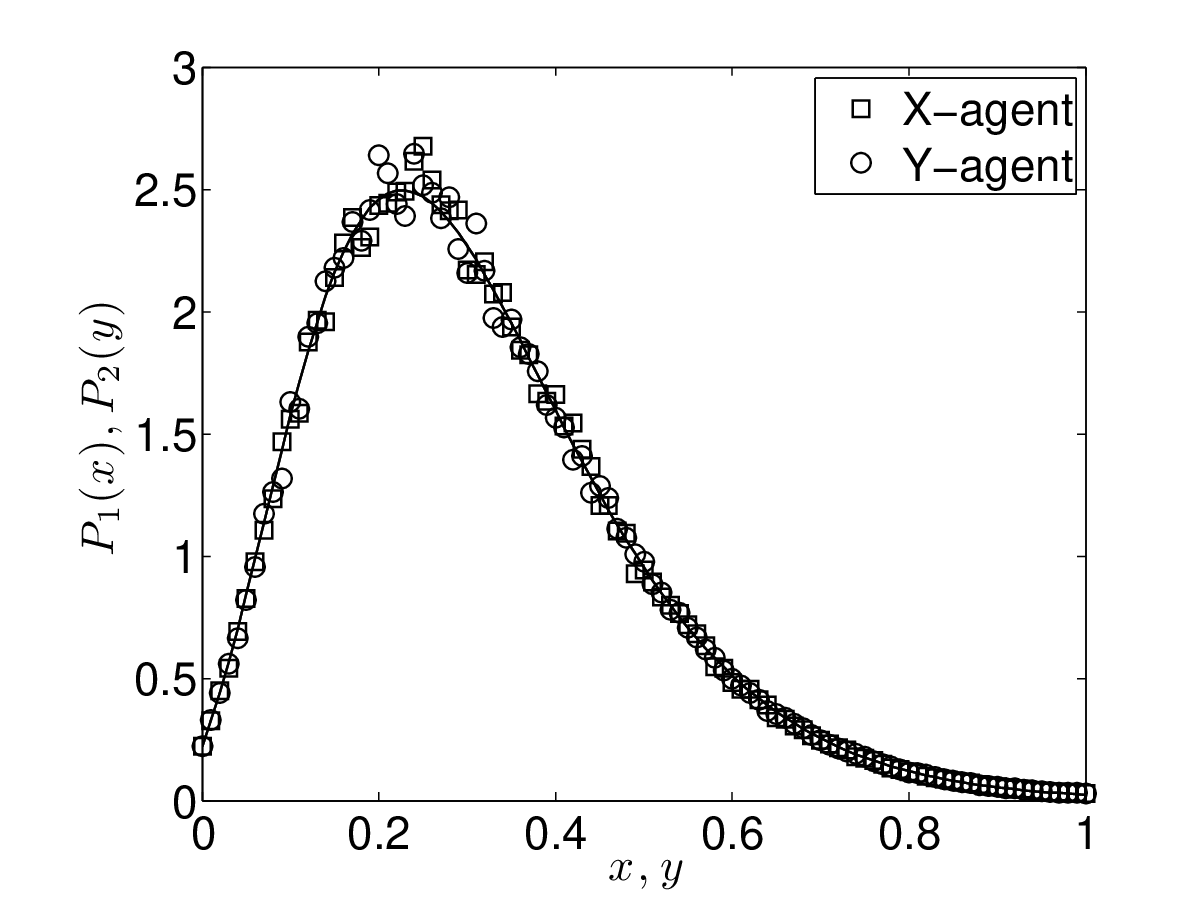}\label{fig:polit8a}
    }
    \subfigure[]{
    \includegraphics[width=0.35\textwidth]{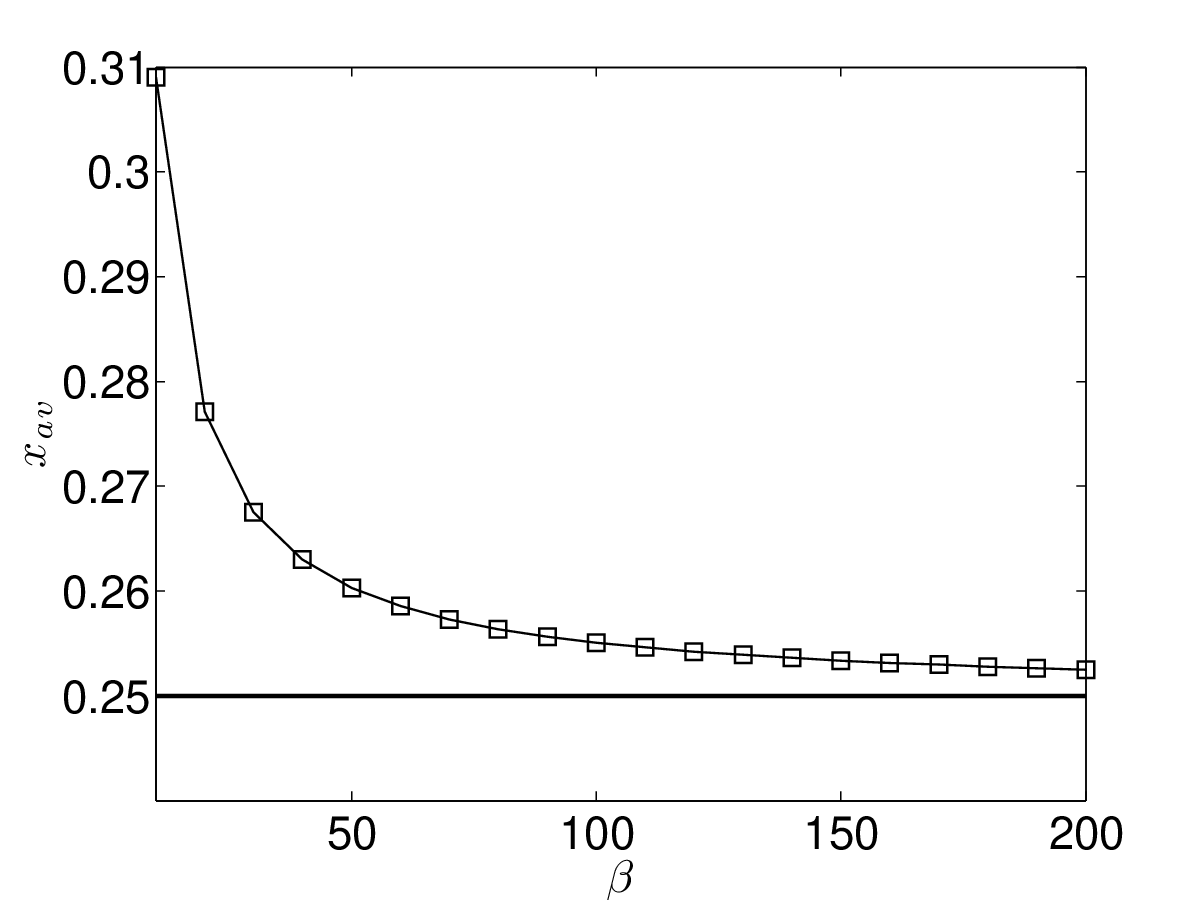} \label{fig:polit8b}
    }
       \caption{(a) Steady--state strategy profiles for the political  advertisement game. The solid line is the prediction of the replicator model, and the symbols are the results of simulations; (b) Average strategy (expenditure) of an agent plotted against the inverse temperature $\beta$. The horizontal line corresponds to the Nash equilibrium $x^*=1/4$.  }
     \label{fig:2}
\end{figure}

We studied the above game using both numerical solutions to the replicator dynamics equations, as well as actual simulations of the game.  In Figure~\ref{fig:polit8a} we plot the steady state strategy profile of the agents in the advertisement game. Again, the solution is symmetric. The strategy profile of each agent seems to be centered around the pure Nash equilbrium, with the exact shape of the density depending on the parameter $\beta$. The average strategy of an agent, defined as $x_{avg}=\int dx P_1(x)x$ is close, but not equal to the Nash equilibrium point $x^*$. However, the strategy average tends asymptotically to this value as one increases the parameter $\beta$, as demonstrated in Figure~\ref{fig:polit8b}. 

\subsection{Investment Game}
\label{sec:investment}
\begin{figure}[!h]
\center
 \includegraphics[width=0.35\textwidth]{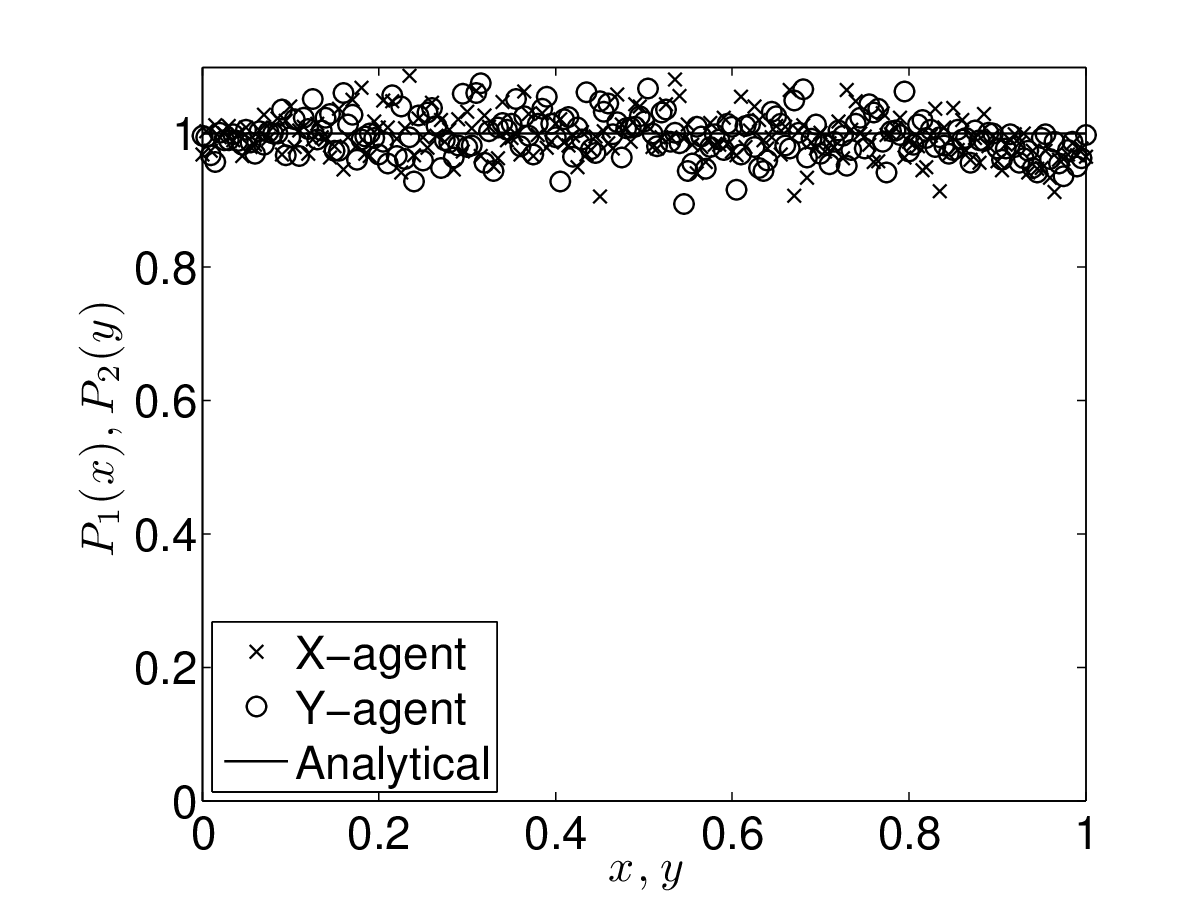}\\
\caption{ Steady--state strategy profiles for the Investment Game. }
\label{fig:election9} 
\end{figure}

Our last example is  a game for which the steady state of the replicator dynamics coincides with the Nash Equilibrium. Consider a model of two--firm competition where each firm chooses an investment levels  from the unit interval.  The firm with the highest investment wins the market, which has unit value, with ties broken randomly. Denoting the investment levels of the players as $x$ and $y$, the payoff structure has the following form: 
\begin{eqnarray}
f_X(x,y) = \left\{ \begin{array}{ll}
         1-x & \mbox{if $x > y$};\\
           -x & \mbox{if $x<y$};\\
        \frac{1}{2}-x & \mbox{if $x =y$}.
        \end{array} \right.
\end{eqnarray}
with a similar (symmetric) payoff for the second agent. Note that the game does not have a pure Nash Equilibrium. Indeed, assume the contrary, and let $x^*$ be the equilibrium strategy of the first agent.  If $x^*<1$, then the second agent will do better by playing $y =  x^* + \varepsilon <1$, while if $x^*=1$, then he will be better off playing $y=0$. At the same time, one can show that there is a  Nash equilibrium where both players mix uniformly over  the pure strategies. It is straightforward to show that the same strategy profiles are also the steady state solutions of the replicator equation. Indeed, note that for a strategy profile $P(y)$ of the second agent, the average reward of the first one is given as 
\begin{equation}
R_1(x) = -x + \int_0^x P(y),
\end{equation}
Then a simple inspection shows  that  choosing $P(x)=P(y)=1$ solves the steady state equations. This is shown in Figure~\ref{fig:election9}, where we also show the simulation results.

Note that   there is an intuitive argument why the steady state of the replicator dynamics corresponds to the Nash equilibrium for  this particular game.  Indeed, recall that the Nash equilibrium minimizes the energy, while the replicator dynamics minimizes the free energy functional~\ref{eq:functional}, which, at non--zero temperature, is different from the energy.  Those two minimization objectives will generally yield different strategy profiles. For this particular game, however, the strategy profile that minimizes the energy also happens to maximize the entropy (and thus, minimize the entropic term in the free energy), so that minimization of either objective yields the same result.   

\section{Conclusion}
\label{sec:discussion}
In this paper we presented a generalization of the replicator dynamics framework to study the dynamics of multi--agent reinforcement learning with  continuous strategy spaces. We presented a set of differential--functional equations that describe the adaptive learning dynamics in multi--agent settings. We also derived a set of equations that characterize the steady state strategy profile of the learning agents. We demonstrated the analytical framework  on several examples, and obtained an excellent agreement with the simulations results. 

It was shown, both theoretically and through simulations, that for the Boltzmann exploration mechanism, the long--term limit of the replicator dynamics does not necessarily correspond to the Nash equilibria of the corresponding game--theoretical system. The reason for this is the bounded rationality of the agents as characterized by the exploration noise in their strategies. Specifically,   the replicator dynamics at non--zero temperature minimizes  an objective function that is related to the thermodynamic concept of  {\em free energy}.  We demonstrated on  examples that the Nash equilibrium is  recovered from the replicator dynamics in the zero temperature limit $\beta \rightarrow \infty$.  Finally, note that for the investment  game considered in Section~\ref{sec:investment}, the steady state of the replicator dynamics coincides with the Nash equilibria for arbitrary $\beta$: The underlying reason for this is the strategy profile that minimizes the energy (NE) also happens to maximize the entropy, so that the the same strategy profile minimizes both energy and free energy. More generally, one can say that any uniform Nash equilibrium also serves as a steady state for the replicator dynamics. 

There are several important directions to pursue this work further. First of all, it will be worthwhile to perform more formal analysis of the steady state equations, and examine the issues of existence and uniqueness of the solutions depending on particular payoff structures. Furthermore, we intend to extend our analysis beyond  two--player games, and specifically, consider games with a very large number of agents, $N\gg1$, where statistical--mechanical approaches for analyzing the solution structure might be appropriate. Finally, another direction of further research is to generalize the stateless Q--learning model considered here to a more general model which will account for different states and probabilistic transitions between them. We believe that such generalization will be possible by complementing the replicator equations with equations that describe the Markovian evolution of the states. 

\section{Acknowledgments}
We thank Jim Crutchfield for useful discussions leading to this work. The work presented here  was supported in part by the USC ISI Intelligent Systems Division's Small Research Grant.

\section*{Appendix}

Recall that the steady state solution of the replicator equations for the bi--linear games is given by Equations~\ref{eq:linear_steady}, where $\gamma_1=\gamma_2\equiv \gamma$ is to be found from Equations~\ref{eq:selfconsist} and ~\ref{eq:g}. Consider the case $a>0$, $b=-a/2$. It is easy to see from Equation~\ref{eq:g} that $g(\gamma)$ nullifies at the origin, so that $\gamma=0$ is the solution of Equation~\ref{eq:selfconsist}; see Figure~\ref{fig:Lin1a}. For sufficiently small $\beta$, this is the only solution. However, there is a critical value $\beta_c$ such that for $\beta > \beta_c$ two more solutions appear. This critical value can be found by equating the derivatives of both sides of Equation~\ref{eq:selfconsist} at $\gamma=0$, which yields $\beta_c =  [g^{\prime}(0)]^{-1}=12/a$.      

Further, we note that the solution $\gamma = 0$ corresponds to a uniform steady state profile, $P(x)=1, x \in [0,1]$.  To examine the stability of this solution, we introduce a small perturbation $\varepsilon(x,t)$ so that  $P(x) \rightarrow P(x) + \varepsilon(x,t)$. Note that relevant perturbations should satisfy  $\int dx \varepsilon(x,t) \equiv 0$ due to normalization. We next linearize the replicator Equation~\ref{eq:rep} around the uniform steady state profile, and  obtain after straightforward calculations
\begin{equation}
\frac{d}{dt}\varepsilon(x,t) =\biggl ( ax-\frac{a}{2} \biggr ) \int dx x \varepsilon(x,t) - T \varepsilon(x,t)
\label{eq:perturbation1}
\end{equation}
Next, let us define $\delta  = \int dx x \varepsilon(x,t)$, which describes  the shift in the mean due to the perturbation. We multiply Equation~\ref{eq:perturbation1} by $x$ and integrate over $x$ to obtain 
\begin{equation}
\frac{d\delta }{dt}  =\biggl ( \frac{a}{12}  - T \biggr ) \delta 
\label{eq:perturbation2}
\end{equation}
Thus, for $T<a/12$, or alternatively, $\beta>12/a\equiv \beta_c$, any perturbation that shifts the mean of the uniform solution will be amplified, indicating that the uniform steady state profile  is unstable.


\begin{thebibliography}{}

\bibitem{abdallah2008} Abdallah, S., and Lesser, V., ``A multiagent reinforcement learning algorithm with non-linear dynamics", Journal
of Artificial Intelligence Research 33:521Ð549, 2008.

\bibitem{Busoniu2008} Busoniu, L.; Babuska, R.; De Schutter, B., ``A Comprehensive Survey of Multiagent Reinforcement Learning," Systems, Man, and Cybernetics, Part C: Applications and Reviews, IEEE Transactions on , vol.38, no.2, pp.156-172, March 2008. 

\bibitem{Borgers1997}
\newblock Tilman Borgers and Rajiv Sarin.
\newblock Learning through reinforcement and replicator dynamics,.
\newblock {\em J. of Economic Theory}, 77(1):1 -- 14, 1997.

\bibitem{Bowling02multiagentlearning}
Michael Bowling and Manuela Veloso.
\newblock Multiagent learning using a variable learning rate.
\newblock {\em Artificial Intelligence}, 136:215--250, 2002.

\bibitem{Claus1998}
Caroline Claus and Craig Boutilier.
\newblock The dynamics of reinforcement learning in cooperative multiagent
  systems.
\newblock In {\em In Proceedings of the Fifteenth National Conference on
  Artificial Intelligence}, pages 746--752. AAAI Press, 1998.

\bibitem{Cressman2005}
Ross Cressman.
\newblock Stability of the replicator equation with continuous strategy space.
\newblock {\em Math. Social Sciences}, 50(2):127 -- 147, 2005.


\bibitem{Hennes2009} Daniel Hennes, Karl Tuyls, Matthias Rauterberg, ``State-coupled replicator dynamics", AAMAS (2) 2009: 789-796. 

\bibitem{Hofbauer1988} 
\newblock J. Hofbauer and K. Sigmund 
\newblock  The theory of evolution and dynamical systems
\newblock Cambridge Univ. Press, Cambridge, U.K., 1988.



\bibitem{Hu98multiagentreinforcement}
Junling Hu and Michael~P. Wellman.
\newblock Multiagent reinforcement learning: Theoretical framework and an
  algorithm.
\newblock In {\em In Proceedings of the Fifteenth International Conference on
  Machine Learning}, pages 242--250. Morgan Kaufmann, 1998.
  
  \bibitem{Hu2003} Junling Hu and Michael~P. Wellman, ``Nash Q-learning for general-sum stochastic games",  JMLR, 4, 1039Ð1069, 2003. 

\bibitem{Kapetanakis2002}
Spiros Kapetanakis and Daniel Kudenko.
\newblock Reinforcement learning of coordination in cooperative multi-agent
  systems.
\newblock In {\em Eighteenth national conference on Artificial intelligence},
  pages 326--331, Menlo Park, CA, USA, 2002.
  
\bibitem{Kaelbling}
\newblock Kaelbling, Leslie  P.  and Littman, Michael  L.  and Moore, Andrew  P.
\newblock Reinforcement Learning: A Survey. 
\newblock JAIR {\bf 4}, pp. 237--285, 1996
  
   
\bibitem{Killingback1999}  T. Killingback, M. Doebeli and N. Knowlton, ``Variable investment, the continuous prisonner's dilemma, and the origin of cooperation", {\em Proc. Roy. Soc. London B, Biol. Sci.} {\bf 266}, pp. 1723Ð1728, 1999.


\bibitem{Le2007} S. Le and R. Boyd, ``Evolutionary dynamics of the continuous iterated Prisoner's dilemma ", {\em J.  Theor. Biology} {\bf  245}, 2, pp. 258-267, 2007.

\bibitem{Oechssler2001}
Joerg Oechssler and Frank Riedel.
\newblock Evolutionary dynamics on infinite strategy spaces.
\newblock {\em Econ. Theory}, 17:141--162, 2001.

\bibitem{Panait2008}
Liviu Panait, Karl Tuyls, and Sean Luke.
\newblock Theoretical advantages of lenient learners: An evolutionary game
  theoretic perspective.
\newblock {\em J. Mach. Learn. Res.}, 9:423--457, 2008.

\bibitem{Peshkin00learningto}
Leonid Peshkin, Kee eung Kim, Nicolas Meuleau, and Leslie~Pack Kaelbling.
\newblock Learning to cooperate via policy search.
\newblock In {\em In UAI}, pages 489--496. Morgan Kaufmann, 2000.

\bibitem{Ruijgrok2005} M. Ruijgrok and T. W. Ruijgrok, ``Replicator dynamics with mutations for games with a continuous strategy space", {\em http://arxiv.org/abs/nlin/0505032}

\bibitem{Sato2002}
Yuzuru Sato, Eizo Akiyama, and J.~Doyne Farmer.
\newblock {Chaos in learning a simple two-person game}.
\newblock {\em Proceedings of the National Academy of Sciences of the United
  States of America}, 99(7):4748--4751, 2002.

\bibitem{Sato2003}
Yuzuru Sato and James~P. Crutchfield.
\newblock Coupled replicator equations for the dynamics of learning in
  multiagent systems.
\newblock {\em Phys. Rev. E}, 67(1):015206, Jan 2003.

\bibitem{Shoham}
Y.~Shoham, T.~Grenager, and R.~Powers.
\newblock Multi-agent reinforcement learning: A critical survey.
\newblock Web manuscript, 2003.


\bibitem{Stone2005} Peter Stone, Richard S. Sutton, and Gregory Kuhlmann, ``Reinforcement Learning for RoboCup-Soccer Keepaway", Adaptive Behavior, 13(3):165Ð188, 2005.

\bibitem{SuttonBarto}
Richard~S. Sutton and Andrew~G. Barto.
\newblock {\em Reinforcement Learning: An Introduction}.
\newblock MIT Press, Cambridge, MA, 1998.

\bibitem{Vrancx2008} Peter Vrancx, Karl Tuyls, Ronald L. Westra, ``Switching dynamics of multi-agent learning", AAMAS (1) 2008: 307-313

\bibitem{Tuyls2003} Karl Tuyls, Katja Verbeeck, Tom Lenaerts, ``A selection-mutation model for q-learning in multi-agent systems", AAMAS 2003: 693-700.

\bibitem{Tuyls2005} K. Tuyls, P.J.T. Hoen, and B. Vanschoenwinkel, ``An evolutionary dynamical analysis of multi-agent learning in iterated games",  Autonomous Agents and Multi-Agent Systems, {\bf 12}, pp. 115Ð153, 2006 

\bibitem{Tuyls2009} Karl Tuyls and Ronald Westra, ``Replicator Dynamics in Discrete and Continuous Strategy Spaces",  in {\em Multi-Agent Systems: Simulation and Applications}, Editor(s): ÊAdelinde M. Uhrmacher; ÊDanny Weyns 2009.  

\bibitem{Wahl1999a} L.~M.~Wahl and M.~A.~Nowak, ``The Continuous Prisoner's Dilemma: I. Linear Reactive Strategies", {\em  Journal of Theoretical Biology} {\bf 200},  3, , pp. 307-32, 1999.

\bibitem{Wahl1999b} L.~M.~Wahl and M.~A.~Nowak, ``The continuous Prisoner's dilemma: II. linear reactive strategies with noise", {\em  Journal of Theoretical Biology} {\bf 200}, pp. 323Ð338, 1999.

\bibitem{Watkins92}
C. J. C.~H. Watkins and P.~Dayan.
\newblock Technical note: Q-learning.
\newblock PhD thesis, 1992.

\end{thebibliography}
\end{document}